\documentclass{bmvc2k}

\title{ControlDreamer: Blending Geometry and Style in Text-to-3D}
\addauthor{Yeongtak Oh$^*$}{dualism9306@snu.ac.kr}{1}
\addauthor{Jooyoung Choi$^*$}{jy_choi@snu.ac.kr}{1}
\addauthor{Yongsung Kim}{libary753@snu.ac.kr}{2}
\addauthor{Minjun Park}{minjunpark@snu.ac.kr}{2}
\addauthor{Chaehun Shin}{chaehuny@snu.ac.kr}{1}
\addauthor{Sungroh Yoon$^{\dagger}$}{sryoon@snu.ac.kr}{1,2}

\addinstitution{
Department of Electrical and \\
Computer Engineering, \\
Seoul National University, \\
Seoul, South Korea
}
\addinstitution{
Interdisciplinary Program in \\
Artificial Intelligence, \\
Seoul National University, \\
Seoul, South Korea 
}

\usepackage{kotex}
\usepackage{tikz}
\usepackage{comment}
\usepackage{color}
\usepackage{booktabs} % for professional tables
\usepackage{pifont}
\usepackage{hyperref}
\usepackage[accsupp]{axessibility}  % Improves PDF readability for those with disabilities.
\usepackage{amsmath,amssymb} % define this before the line numbering.
\usepackage{comment}

\hypersetup{colorlinks, urlcolor={medium-blue}}
\hypersetup{
    colorlinks=true,
    linkcolor=purple,
    filecolor=magenta,      
    urlcolor=cyan,
}
\newcommand{\cmark}{\ding{51}}%
\newcommand{\xmark}{\ding{55}}%
\usepackage{tcolorbox}
\tcbuselibrary{xparse}

\NewTColorBox[auto counter, number within=section]{mybox}{o}
  {
   title={Human Evaluation Template},
  }

\runninghead{Oh and Choi \etal}{ControlDreamer}

\def\etal{\emph{et al}\bmvaOneDot}

\begin{document}

\maketitle
\def\thefootnote{*}\footnotetext{These authors contributed equally to this work}
\def\thefootnote{$\dagger$}\footnotetext{Corresponding author}
\begin{abstract}
Recent advancements in text-to-3D generation have significantly contributed to the automation and democratization of 3D content creation. Building upon these developments, we aim to address the limitations of current methods in blending geometries and styles in text-to-3D generation. We introduce multi-view ControlNet, a novel depth-aware multi-view diffusion model trained on generated datasets from a carefully curated text corpus. Our multi-view ControlNet is then integrated into our two-stage pipeline, ControlDreamer, enabling text-guided generation of stylized 3D models. Additionally, we present a comprehensive benchmark for 3D style editing, encompassing a broad range of subjects, including objects, animals, and characters, to further facilitate research on diverse 3D generation. Our comparative analysis reveals that this new pipeline outperforms existing text-to-3D methods as evidenced by human evaluations and CLIP score metrics. Project page:  \href{https://controldreamer.github.io}{https://controldreamer.github.io}
\end{abstract}

\section{Introduction}
3D content creation has recently been gathering attention, particularly in the realms of virtual reality and game development. This evolving landscape has been profoundly impacted by the recent advancement of 2D lifting methods. Dreamfusion's~\cite{poole2022dreamfusion} introduction of score distillation sampling (SDS) has revolutionized and democratized the field, facilitating the generation of 3D models by leveraging large-scale text-to-image diffusion models~\cite{saharia2022photorealistic,rombach2022high,balaji2022ediffi}. This breakthrough has enabled more intuitive and imaginative 3D content creation from users' textual descriptions. Furthering this evolution, various works show high-resolution text-to-3D generation, employing multiple 3D representations~\cite{lin2023magic3d,chen2023text2tex,chen2023fantasia3d} or variants of score distillation sampling~\cite{wang2023prolificdreamer,haque2023instruct}.
One such approach is MVDream \cite{shi2023mvdream}, a model that extends the capabilities of Stable Diffusion \cite{rombach2022high}, a well-known text-to-image diffusion model, by adapting it on the Objaverse dataset \cite{deitke2023objaverse}, which consists of multi-view images of 3D objects.

\begin{figure*}[t!]
\centering
\includegraphics[width=0.9\linewidth]{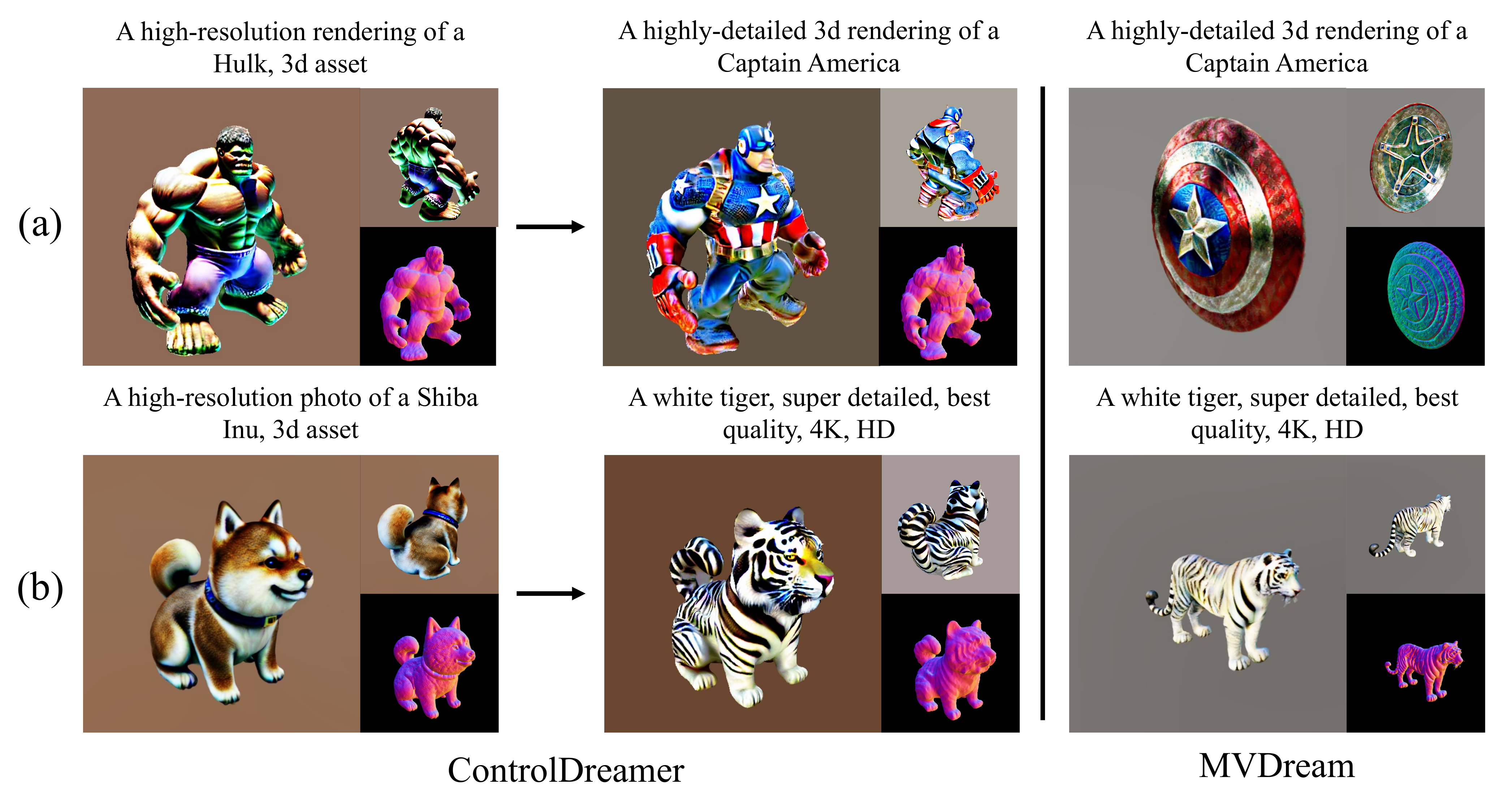} 
\caption{Comparing text-to-3D pipeline of our ControlDreamer and MVDream. On the right, MVDream’s output reveals a vulnerability to pre-training geometry biases, often producing (a) unintended results such as shields or (b) stereotypical geometries related to prompts. On the left, ControlDreamer overcomes these biases, enabling unique combinations of geometry and style, even facilitating counterfactual generation in 3D models.}
\label{fig:fig1}
\vspace{-1em}
\end{figure*}

However, we observe that when attempting to generate imaginative 3D models using MVDream, it fails to seamlessly blend attributes such as geometry and style. For instance, when given a prompt like `Captain America,' the model tends to produce a 3D model of a shield rather than the character itself. This issue likely arises due to dataset biases and the mode-seeking nature of score distillation sampling \cite{poole2022dreamfusion}, which causes the model to generate geometries that reflect dominant modes in the dataset, rather than customized or more nuanced 3D models, such as a bulky version of Captain America (Fig. \ref{fig:fig1}).

%%%%% Figure  1 %%%%%%%%%%%%%%%%%%
%%%%%%%%%%%%%%%%%%%%%%%%%%%%%%
These challenges motivated us to decompose the text-to-3D pipeline into a two-stage approach given two text prompts. Initially, we generate a neural radiance field (NeRF)~\cite{mildenhall2021nerf,mueller2022instant} from a \textit{geometry prompt} to sculpt geometric structures, followed by converting it to a mesh using DMTet algorithm~\cite{shen2021deep} and refining both geometry and texture using a \textit{style prompt}. To address this, we present \textit{MV-ControlNet}, a depth-aware multi-view diffusion model. We first generate a paired dataset of text and multi-view images using GPT-4~\cite{gpt4} filtered texts and MVDream, followed by training a depth conditioner for the frozen MVDream, according to the principle of ControlNet~\cite{zhang2023adding}. %Obtaining MV-ControlNet is cost-effective, requiring only 1.5 days on dataset generation and 1.5 days on training, on four NVIDIA A40 GPUs.

In evaluations on our new benchmark, which consists of pairs of geometry and style prompts, ControlDreamer shows superior performance in generating styles on 3D models compared to existing two-stage pipelines. By using CLIP~\cite{radford2021learning} scores and human assessments, we find that our depth-aware MV-ControlNet outperforms normal and edge-aware variants in rendering detailed 3D models. These findings confirm ControlDreamer's ability to produce high-quality 3D models that closely match the provided textual descriptions. 
\begin{table*}[h!]
\centering
\caption{Comparison to previous text-to-3D generation methods. Our method is the first two-stage pipeline designed for multi-view consistent 3D generation. Notably, it offers convenience by eliminating the need for annotations like bounding boxes and masks.}
\label{tab:metadata}
{\resizebox{0.7\textwidth}{!}
{\begin{tabular}{ccccc}
\hline
    \toprule
    Method & 3D Rep. & Multi-Stage & Annot.-Free & Multi-View \\
    \midrule
    ProlificDreamer \cite{wang2023prolificdreamer} & NeRF & \textcolor{red}{\xmark} & \textcolor{green}{\cmark} & \textcolor{red}{\xmark}\\
    Fantasia3D \cite{chen2023fantasia3d} & DMTet & \textcolor{green}{\cmark} & \textcolor{green}{\cmark} & \textcolor{red}{\xmark}\\
    Magic3D \cite{lin2023magic3d} & NeRF$\&$DMTet & \textcolor{green}{\cmark} & \textcolor{green}{\cmark} & \textcolor{red}{\xmark} \\
    MVDream \cite{shi2023mvdream} & NeRF & \textcolor{red}{\xmark} & \textcolor{green}{\cmark} & \textcolor{green}{\cmark}\\
    \midrule
    Ours & NeRF$\&$DMTet & \textcolor{green}{\cmark} & \textcolor{green}{\cmark} & \textcolor{green}{\cmark} \\
    \bottomrule
\end{tabular}}
}
% Vox-E \cite{sella2023vox} & Voxel & \textcolor{green}{\cmark} & \textcolor{red}{\xmark} & \textcolor{red}{\xmark} \\
%     PG3D \cite{cheng2023progressive3d} & NeRF & \textcolor{green}{\cmark} & \textcolor{red}{\xmark} & \textcolor{red}{\xmark}\\
%     TEXTure \cite{richardson2023texture} & Texture & \textcolor{red}{\xmark} & \textcolor{green}{\cmark} & \textcolor{red}{\xmark}\\
%     Text2tex \cite{chen2023text2tex} & Texture & \textcolor{red}{\xmark} & \textcolor{green}{\cmark}& \textcolor{red}{\xmark} \\
%}
\vspace{-1em}
\end{table*}
\section{Related Works}
\subsection{Adding Conditions to Text-to-Image Models}
Diffusion models like Stable Diffusion \cite{rombach2022high}, trained on billions of image-text pairs~\cite{schuhmann2022laion}, have shown remarkable capabilities in generating high-quality images from text or in removing harmful degradations from images~\cite{oh2024efficient}. ControlNet \cite{zhang2023adding} introduces a method for adding conditions to Stable Diffusion by using a trainable copy alongside the original frozen model to perform specific tasks, such as depth-to-image conversion. 
% Imagen \cite{saharia2022photorealistic}, and eDiff-I \cite{balaji2022ediffi}, 
% Instruct-Pix2Pix \cite{brooks2023instructpix2pix} uses GPT-3 \cite{brown2020language} and Stable Diffusion to create a training dataset for training an image editing model that follows diverse editing instructions.
%%%%%%%%%%%%%%%%%%%%%%
%%%%%%%%%%%%%%%%%%%%%%
\subsection{Text-to-3D Generation}
%\textcolor{blue}{Despite recent advancements in 3D-aware generative models~\cite{voleti2024sv3d, boss2024sf3d} trained on large-scale 3D datasets or videos, the challenge of 3D stylization remains significant. 
While public datasets of 2D images are abundant, dataset of 3D assets are relatively scarce ($\sim$80K in Objaverse~\cite{deitke2023objaverse}). Consequently, leveraging pre-trained 2D diffusion models to create diverse, user-customized 3D models continues to be a demanding task. To overcome such difficulties, DreamFusion~\cite{poole2022dreamfusion} introduces score distillation sampling (SDS) to enable the optimization of an implicit 3D model~\cite{mildenhall2021nerf,mueller2022instant} without any 3D dataset, utilizing a pre-trained 2D diffusion model as a prior.
Despite these advantages, DreamFusion suffers from slow scene generation and lower-quality results.
To overcome these limitations, several methods are proposed involving primary optimization for NeRF and secondary optimization for the extracted mesh~\cite{lin2023magic3d,chen2023fantasia3d}, along with improvements for SDS~\cite{wang2023prolificdreamer, katzir2023nfsd,metzer2023latent, chen2024sculpt3d}. In contrast to prior 2D lifting-based methodologies, MVDream~\cite{shi2023mvdream} fine-tunes a text-to-image diffusion model~\cite{rombach2022high} through joint training on large-scale 2D and 3D datasets. Consequently, MVDream circumvents the \textit{Janus Problem} and the \textit{Content Drifting Problem}, consistently generating 3D models that align accurately with the given texts.
Concurrent works have developed large-scale 3D generative models~\cite{voleti2024sv3d, boss2024sf3d}. However, these models primarily rely on image inputs, making them unsuitable for our objective of controlling both the geometry and style of 3D models through text descriptions.
\begin{figure*}[t!]
\centering
\includegraphics[width=0.9\linewidth]{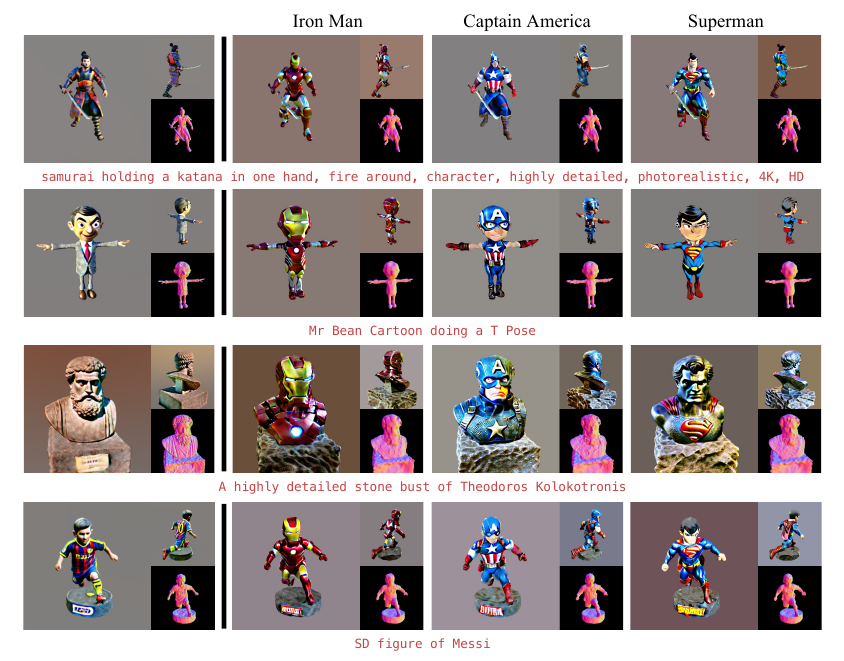} 
\caption{Main results. Our generation process begins by generating a coarse-grained geometry, followed by creating a fine-grained stylized 3D model using a style prompt.
}
\label{fig:multi_sources}
\end{figure*}
% \textcolor{blue}{Sculpt3D, SV3D, IT3D}
%%%%%%%
% \vspace{-0.5em}
\subsection{Text-Guided 3D Editing}
In text-to-3D generation, various innovative methods have been proposed. 
% Vox-E~\cite{sella2023vox} edits voxel grids with SDS, maintaining geometry and appearance. Instruct-NeRF2NeRF~\cite{haque2023instruct} iteratively updates image dataset using Instruct-Pix2Pix~\cite{brooks2023instructpix2pix}. Text2tex~\cite{chen2023text2tex} assumes a perfect mesh, applying textures without mesh refinement flexibility. 
Magic3D~\cite{lin2023magic3d} offers a coarse-to-fine pipeline for high-resolution 3D models, starting with NeRF and converting to deformable meshes. Fantasia3D~\cite{chen2023fantasia3d} divides its process into geometry and appearance stages, beginning with a 3D ellipsoid and refining with DMTet. 
% Progressive3D~\cite{cheng2023progressive3d} decomposes complex prompts for localized editing, focusing on user-specified regions.
TEXTure~\cite{richardson2023texture} handles a similar task of generating textures on top of geometry, but unlike our approach, it does not allow for changes in the geometry. IT3D~\cite{chen2024it3d} refines initial rendered views using a 2D ControlNet~\cite{zhang2023adding} and ensures multi-view consistency through a GAN loss, whereas we use a single multi-view diffusion model.
In Table \ref{tab:metadata}, we outline a key gap in existing research: a lack of methods combining multi-stage, annotation-free, and multi-view capabilities for 3D generation. We found that current methods, especially in DMTet texture refinement, work well only when geometry and style prompts match. Furthermore, methods using 2D models struggle with multi-view consistency. Our ControlDreamer generates multi-view consistent 3D models by combining geometry and style specified by text prompts. 
%\textcolor{blue}{Moreover, unlike previous methods that cannot alter the source geometry during the refinement stage~\cite{richardson2023texture}, our approach enables plausible geometry and texture editing based on creative, user-customized target text (see Fig.~\ref{fig:multi_sources}). IT3D~\cite{chen2024it3d} 고민}
% TEXTure~\cite{}}
% \vspace{-1.5em}
%%%%%
%Addressing this, we integrated multi-view features from MVDream in our ControlDreamer, enabling high-quality, differentiated prompt-based 3D generation.
\begin{figure*}[t!]
\centering
\includegraphics[width=0.9\linewidth]{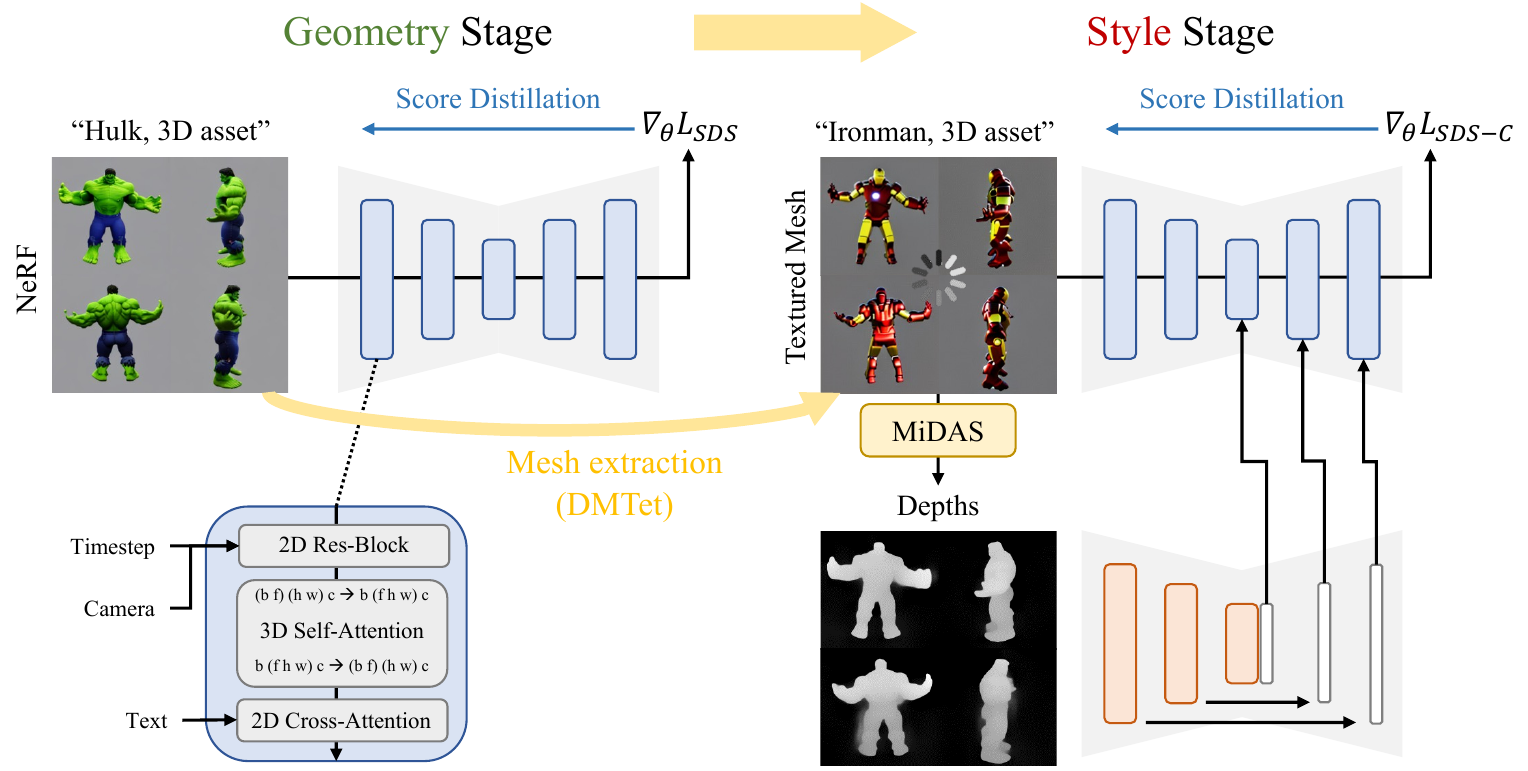} 
\caption{Illustration of ControlDreamer. (Left) Starting with a geometry prompt, we use MVDream to generate a NeRF, ensuring consistency through 3D self-attention. (Right) The NeRF is converted into a mesh via DMTet, followed by style generation through our MV-ControlNet, which integrates a trainable copy (red) and employs zero-initialized convolutions (white). MV-ControlNet is designed to understand geometry using multi-view depth.}
\label{fig:main}
\end{figure*}
\section{Method}
%%%%%
Our ControlDreamer pipeline involves a two-stage process: first, training a coarse-grained NeRF~\cite{mildenhall2021nerf,mueller2022instant} with a geometry prompt %and converting it into a textured mesh using DMTet, 
and then creating a fine-grained stylized textured mesh~\cite{shen2021deep} with a style prompt. We illustrate our pipeline in Fig. \ref{fig:main} and provide details in Sec. \ref{sec:3.1}. Then, we emphasize a depth-aware, multi-view approach in the second stage and elaborate on the Multi-View (MV) ControlNet's training scheme in Sec. \ref{sec:3.2}.
%%%%%%
\subsection{ControlDreamer: Two-Stage 3D Generation}\label{sec:3.1}
\subsubsection{Geometry Stage}
Here, we generate a NeRF from a given geometry prompt. The style of this NeRF is subsequently modified in the second stage. We optimize the NeRF by distilling probability density~\cite{oord2018parallel} from the pre-trained MVDream \cite{shi2023mvdream}. This optimization is achieved through score distillation sampling (SDS)~\cite{poole2022dreamfusion}, where the diffusion model directly computes the loss gradient as shown in the following equation:
% \vspace{-0.5em}
\begin{equation}\label{eq:sds}
\nabla_{\theta} \mathcal{L}_{\text{SDS}} \triangleq \mathbb{E}_{t, \epsilon} \left[ w(t) \left( \hat{\epsilon}_{\phi}(z_t; y_{\text{geo}}, c, t) - \epsilon \right) \frac{\partial z}{\partial x} \frac{\partial x}{\partial \theta} \right]
,
\end{equation}
where $t \sim U(t_{\text{min}},t_{\text{max}})$, $w(t)$ denotes the weighting function, $\hat{\epsilon}_{\phi}(z_t; y, t)$ represents a guided epsilon prediction \cite{ho2022classifier}, $y_{\text{geo}}$ is the geometry text prompt, $c$ represents the camera parameters, $z$ refers to the VAE latents \cite{kingma2013auto,rombach2022high} of rendered multi-view images $x$, and $\theta$ is the parameter of 3D model. Note that there is no backpropagation through the diffusion model.

To leverage the multi-view consistency of MVDream, we render four views using randomly selected camera parameters $c$ in each iteration of optimization. These parameters are uniformly sampled from a range of view angles, each maintaining consistent elevation. 
In further detail, we incorporate point lighting \cite{poole2022dreamfusion}, orientation loss \cite{poole2022dreamfusion}, and soft shading \cite{lin2023magic3d}. 
%Additionally, we enhance the background with random color augmentation and employ a fixed negative prompt to avoid low-quality 3D models. 
The timestep range $[t_{\text{min}},t_{\text{max}}]$ is linearly annealed during the optimization process. To prevent color saturation, we apply the $x_0$-reconstruction loss, which is the epsilon reconstruction loss weighted by the signal-to-noise ratio (SNR) \cite{kingma2021variational}, thus $w(t)=\text{SNR}(t)$. %The text prompt $y_{geo}$ is embedded using the CLIP ViT-L text encoder~\cite{radford2021learning}.
%%%%%%%%%%%%
\subsubsection{Style Stage}
Our second stage is motivated by the coarse-to-fine optimization of \cite{lin2023magic3d}, which is utilized for generating high-resolution meshes. In this process, they optimize the textured mesh using DMTet \cite{shen2021deep}, which uses a deformable tetrahedral grid and employs a differentiable rasterizer \cite{Laine2020diffrast, Munkberg2022nvdiffrec} for efficient rendering. However, unlike the goal of \cite{lin2023magic3d}, which is centered on high-resolution mesh refinement, our approach leverages the DMTet algorithm for generating \textit{style} on the previously generated geometry.

In each iteration of optimization, we employ pre-trained MiDAS~\cite{Ranftl2022midas} to obtain predicted depth maps, using the same depth estimator that is used for training our MV-ControlNet in Sec. \ref{sec:3.2}. Subsequently, we optimize the mesh with the gradient computed by MV-ControlNet as shown in the following equation:
\begin{equation}\label{eq:sds_controlnet}
\nabla_{\theta} \mathcal{L}_{\text{SDS}-\text{C}} \triangleq \mathbb{E}_{t, \epsilon} \left[ w(t) \left( \hat{\epsilon}_{\phi}(z_t;\mathcal{D}(x), y_{\text{style}}, c, t) - \epsilon \right) \frac{\partial z}{\partial x} \frac{\partial x}{\partial \theta} \right]
,
\end{equation}
where $y_{\text{style}}$ is the style text prompt, which is used differently from $y_{\text{geo}}$. $\mathcal{D}$ is our MV-ControlNet, which we later explain in Sec. \ref{sec:3.2}. This gradient is coupled with regularizations to refine both geometry and texture meticulously. In contrast to recent methods \cite{lin2023magic3d,chen2023fantasia3d,wang2023prolificdreamer} that necessitate narrow $[t_{\text{min}},t_{\text{max}}]$ during the refinement phase, our ControlNet allows a large range of $[t_{\text{min}},t_{\text{max}}]$ with scheduled timestep annealing. This approach ensures the depth information remains intact maintaining original geometry throughout the style optimization, even with significant timestep variations. 
%We discover that the iterative updating of the DMTet mesh, guided by estimated depth maps, effectively modifies the geometry and texture in the style stage.
%%%%%%%%%%%%
% \subsubsection{Regularizations}
\subsubsection{Training Details}
We use the normal map consistency loss~\cite{lin2023magic3d} and the Laplacian smoothing regularization loss~\cite{wang2023prolificdreamer} to eliminate pixelated artifacts in DMTet. These allow us to generate photorealistic 3D models. Utilizing MV-ControlNet, textures are effectively modified while preserving the initial geometry, even with significant variations in text descriptions. Furthermore, this effectiveness persists in scenarios where $t$ is sampled from $t \sim U(0.02, 0.98)$. In the style stage, we sample $t$ from this range for the initial $4/5$ of the total iterations, akin to the geometry stage. Subsequently, for the remaining iterations, $t$ is sampled from $t \sim U(0.02, 0.5)$. These techniques differentiate MV-ControlNet from previous methods~\cite{lin2023magic3d,chen2023fantasia3d} that rely on smaller timesteps for mesh refinement.

\subsubsection{Theoretical Discussion}
The key to the success of score distillation in text-to-3D generation lies in the guidance mechanism~\cite{ho2022classifier}. The guided score of Eq. \ref{eq:sds} can be written as the following equation:
\begin{align}\label{eq:cfg}
\hat{\epsilon}_{\phi}(z_t;y_{\text{geo}})
&=s\cdot(\epsilon_{\phi}(z_t;y_{\text{geo}})-\epsilon_{\phi}(z_t))+\epsilon_{\phi}(z_t) \\
&\propto \nabla_{z_t}(s\cdot( \text{log}p(z_t|y_{\text{geo}}) - \text{log}p(z_t)) + \text{log}p(z_t)) \\
&=\nabla_{z_t} ( s\cdot\text{log}p(y_{\text{geo}}|z_t)+\text{log}p(z_t)),
\end{align}
due to Bayes' rule. Here, $s$ represents the guidance scale and $c$ and $t$ are omitted for brevity. The equation indicates strong guidance encourages higher text alignment $p(y_{\text{geo}}|z_t)$.

The effect of integrating our MV-ControlNet into score distillation can be demonstrated similarly. The guided score in Eq. \ref{eq:sds_controlnet} can be expressed as follows:
\begin{align}\label{eq:cfg-c}
    \hat{\epsilon}_{\phi}(z_t;\mathcal{D}(x),y_{\text{style}})
    &\propto \nabla_{z_t} ( s\cdot(\text{log}p(z_t|\mathcal{D}(x), y_{\text{style}})-\text{log}p(z_t|\mathcal{D}(x))) +  \text{log}p(z_t|\mathcal{D}(x))) \\
    &= \nabla_{z_t} ( s\cdot\text{log}\underbrace{p(y_{\text{style}}|z_t,\mathcal{D}(x))}_{\text{style blending}} +  \text{log}\underbrace{p(z_t|\mathcal{D}(x))}_{\text{geometry}}),
\end{align}
also due to Bayes' rule. Here, maximizing $p(z_t|\mathcal{D}(x))$ ensure that the optimized 3D representation maintains the geometry, while maximizing $p(y_{\text{style}}|z_t,\mathcal{D}(x))$ guides the blending of a style that suits the geometry.
%%%%%%%%%%
% \subsubsection{Annealing diffusion timesteps}

% \subsubsection{Editing the geometry and texture}
% ProlificDreamer \cite{wang2023prolificdreamer} comprises two refinement stages: geometric refinement followed by texture refinement. Furthermore, each training iteration requires not only the computation for SDS but also the fine-tuning of the diffusion model, resulting in a computationally expensive generation process. Additionally, the text prompts remain unchanged throughout each refinement stage, which restricts the diverse combination of the prompts. However, our method significantly relaxes the stringent requirements of text alignment in the stylization phase by enabling style editing for any 3D model, transforming it into a DMTet mesh format. Our approach offers greater flexibility and creative potential in 3D model stylization.
%%%%%%
\subsection{Multi-View ControlNet}\label{sec:3.2}
As elaborated in the previous section, our pipeline generates style in the second stage by refining the DMTet, converted from NeRF. We note the limitations of MVDream in refining DMTet, stemming from a lack of understanding of the geometry. To address this, we introduce the Multi-View (MV) ControlNet, crucial for generating fine-grained style by understanding coarse-grained geometry.
%%%%%%
\subsubsection{Generating Dataset}
To train a depth-aware multi-view model, we construct a dataset using the curated 100K high-quality texts from OpenShape~\cite{liu2023openshape}, filtered with GPT-4~\cite{gpt4} for quality assurance. Initially, these texts are refined using BLIP~\cite{li2022blip} and Azure Cognition Services for 2D image captions and subsequently enriched by GPT-4 to filter out uninformative content. Using MVDream, we generate multi-view images from these texts, matching MVDream's camera parameters to maintain its prior knowledge. This involves generating four orthogonal views for each text, each with uniformly distributed azimuth angles and elevation. Specifically, we randomly sample the camera parameters in a range of [0.9, 1.1] for distance, [15, 60] for fov, and [0, 30] for elevation. Then, we generate corresponding depth maps from the generated images using MiDAS~\cite{Ranftl2022midas}. This process, completed in 1.5 days on four A40 GPUs, involves texts that either explicitly mention `3D' or have `3D asset' appended as a postfix.
%%%%%%
\subsubsection{Training MV-ControlNet}
%MVDream \cite{shi2023mvdream} expands the text-to-image capabilities of Stable Diffusion~\cite{rombach2022high} to enable text-to-multi-view generation. The model processes inputs of noisy latents $z_t \in \mathbb{R}^{F \times H \times W \times C}$, camera parameters $c \in \mathbb{R}^{F \times 16}$, and text prompts. MVDream differentiates itself by utilizing a 3D self-attention mechanism, an extension of 2D self-attention, to model multi-view consistency and capture the interrelations among pixels across $F$ views. Note that additional operations such as cross-attentions and convolutions operate on a frame-wise basis.

Our goal is to train a depth encoder that incorporates multi-view depth conditions into MVDream, leveraging the model's 3D consistency and generation quality. Following the approach of ControlNet~\cite{zhang2023adding}, which involves training an additional encoder atop a frozen diffusion model, we introduce multi-view depths as additional inputs to the pre-trained MVDream. Specifically, a trainable copy of MVDream's U-Net~\cite{ronneberger2015unet} encoder, including 3D attentions, is tasked with encoding depths 
and integrated into the U-Net decoder using zero-initialized convolutions. 
We train this encoder with the standard diffusion objective~\cite{ho2020denoising}
formalized as follows: 
\begin{equation}\label{eq:controlnet}
\mathcal{L} = \mathbb{E}_{z \sim \mathcal{E}(x),t,\epsilon \sim \mathcal{N}(0,1)} \left[ \left\| \epsilon - \epsilon_{\phi}(z_t;\mathcal{D}(x),y,c,t) \right\|^2 \right],
\end{equation}
where $\mathcal{E}$ denotes the frozen VAE encoder~\cite{rombach2022high}, $\mathcal{D}$ is the trainable depth encoder, $\epsilon_{\phi}$ is the frozen MVDream's U-Net, $y$ is the text prompt, and $c$ is the camera parameter.

% \vspace{-3em}
%%%%%% Figure 4 %%%%%%

%%%%%%%%%%%%%%%%%%%%%%%%%%%%%%%%%%%%%%%%
\begin{figure}[t!]
    \includegraphics[width=0.8\linewidth]{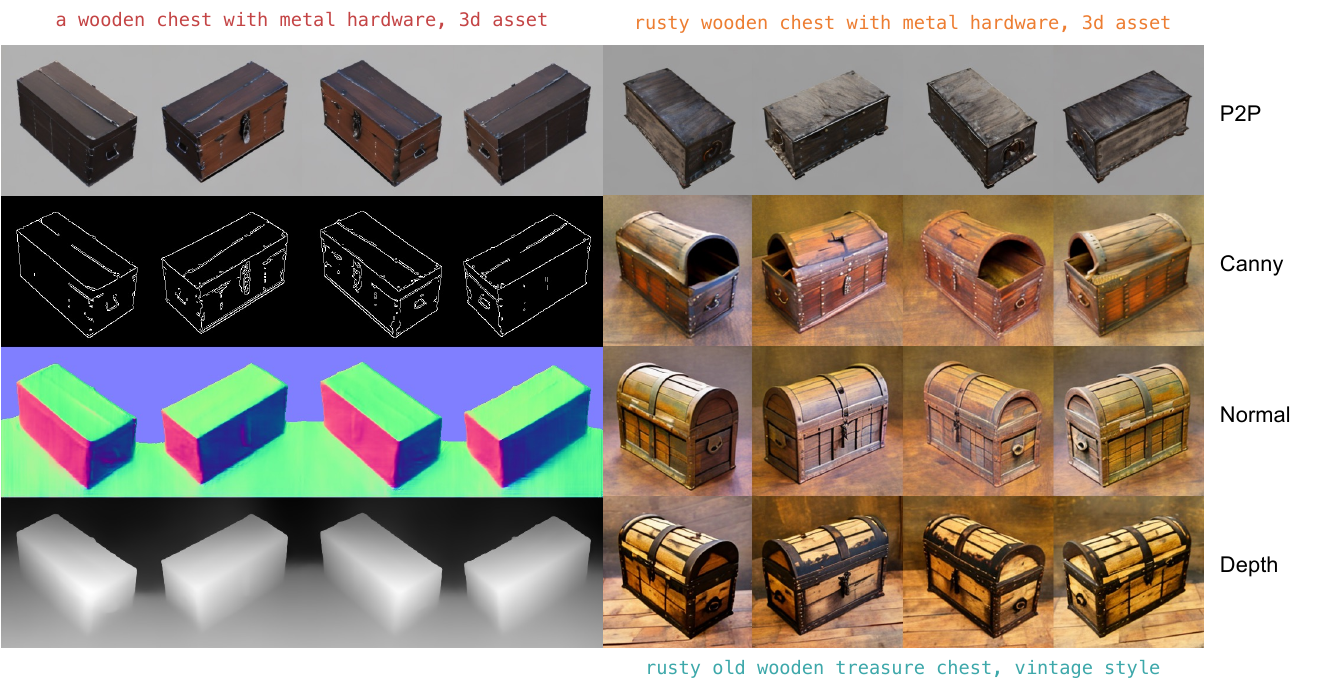} 
    \centering
    \caption{We compare the depth-aware MV-ControlNet with the P2P~\cite{hertz2022prompt} approach on MVDream, and also against MV-ControlNet variants trained under edge and normal conditions. On the left, source images are displayed alongside their respective conditions. Among these, the depth-conditioned multi-view images display the most visually appealing results.}
    \label{fig:control_variants}
\end{figure}

%\begin{figure*}[ht!]
%\centering
%\includegraphics[width=\linewidth]{Figures/style_mixing_draft.pdf} 
%\caption{Samples of our two-stage 3D generation pipeline. Our generation process involves first generating a coarse-grained geometry, and then generating a fine-grained stylized 3D model with a style prompt.}
%\label{fig:style_mixing}
%\end{figure*}
%%%%%%%%%%%%%%%%%%%%%%%%%%%%%%%%%%%%%%%%
\vspace{-1em}
\section{Experiments}
%In our experiments, we carry out tests using user-customized prompts, distinct from the $100K$ text corpus used for training our MV-ControlNet. 
For both qualitative and quantitative assessments, we have created a benchmark dataset comprising a total of 30 pairs of geometry and style prompts. These pairs span across five categories: \textit{Animals, Characters, Foods, General}, and \textit{Objects}. Qualitative samples of our pipeline are presented in Fig. \ref{fig:multi_sources}. % The main qualitative samples of our pipeline are presented in Fig. \ref{fig:multi_sources}.

% - high quality의 text를 사용해서 text corpus를 refine해서 training하는 경우, distill받는 model의 학습 성능이 크게 증가하더라 정도의 학습 비교 필요 \\
% \subsection{ControlNet의 in-domain / out-of-domain text에 대한 control 성능 실험}
%\subsection{Comparative Studies}
%\textcolor{blue}{For the comparison of input conditions—Canny edge, Normal, and Depth—we obtain images corresponding to each condition from a pre-generated multi-view image dataset using $100K$ text corpus. Subsequently, we train the models for each condition to assess the level of controllability on multi-view image generation and 3D stylization.}
%%%%%%
\subsection{Qualitative Comparison of Multi-View Image Generation}
Before evaluating 3D generation, we assess MV-ControlNet's multi-view image generation capabilities by comparing with several baselines as shown in  Fig. \ref{fig:control_variants}. The baselines include MVDream using P2P \cite{hertz2022prompt}, an image editing method, and MV-ControlNet variants trained under the canny edge and normal maps. 
Regarding MV-ControlNet variants, we observe that canny edge maps result in unrealistic and 3D-inconsistent images, whereas normal maps provide 3D consistency but show degraded text alignment. Conversely, using depth conditioning proves most effective, leading to visually appealing and well-aligned results. This success motivated us to choose depth-conditioned MV-ControlNet as our model. %Additional experimental details and results are available in the appendix.
%%%%%
% arxiving 이후!
% \textcolor{blue}{CLIP Score로 MVDream vs MV ControlNet, GPT4로 3D 렌더링의 text-alignmnet 평가.}

%%%%%%%%%%%%%%%%%%%%%%%%%%%%%%%%%%%%%
\begin{figure*}[ht!]
\centering
\includegraphics[width=0.9\linewidth]{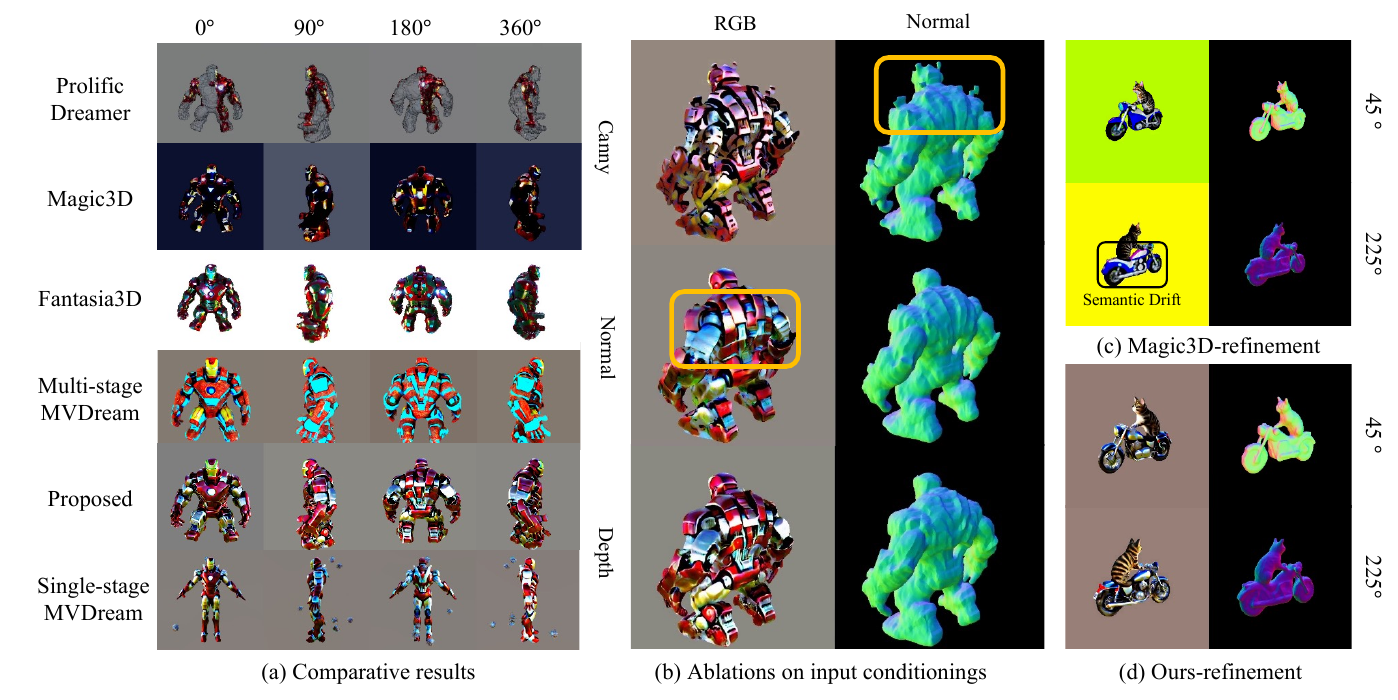} 
\caption{In (a), we present comparisons with previous pipelines. Hulk's geometry from Fig. \ref{fig:fig1}, styled as Ironman, reveals that Magic3D and ProlificDreamer often produce texture artifacts, while Fantasia3D and MVDream are prone to color oversaturation. (b) illustrates the results under various input conditions. (c) shows the refinement process using Magic3D, while (d) highlights the superior results achieved with our ControlDreamer.}
\label{fig:compare}
\end{figure*}
%%%%%%%%%%%%%%%
%%%
\begin{figure}[ht!]
\centering
\includegraphics[width=0.9\linewidth]{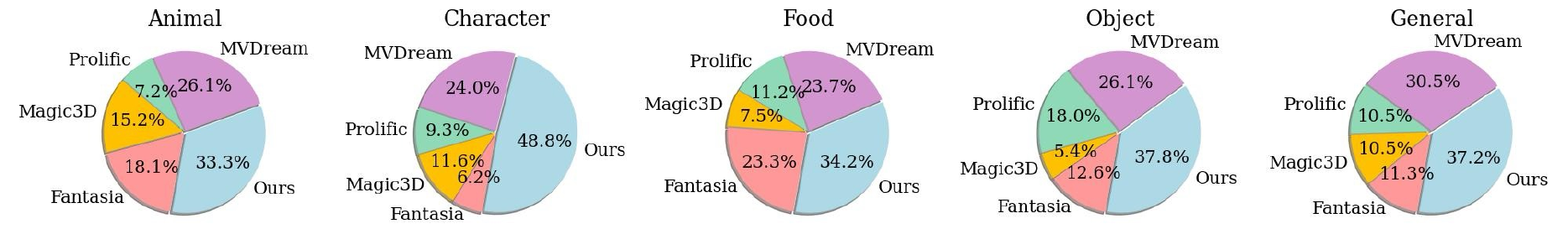} 
\caption{Human preference study on editability and text alignment in five prompt domains. ControlDreamer consistently ranks highest in user preferences across all domains.}
\label{fig:human_eval}
\end{figure}
\subsection{Qualitative Comparison of 3D Generation}
We compare against several baselines, namely Magic3D~\cite{lin2023magic3d}, Fantasia3D~\cite{chen2023fantasia3d}, and ProlificDreamer~\cite{wang2023prolificdreamer}. All these methods, including our ControlDreamer, employ the DMTet~\cite{shen2021deep} algorithm to refine textured meshes. The results of our stylization experiments using these baseline models are illustrated in Fig.~\ref{fig:compare}. With Magic3D and ProlificDreamer, we observe some instances of failure, characterized by only partial texture changes in the geometry. This issue arises from a misalignment between the prior knowledge embedded in Stable Diffusion for the style prompt (e.g., the Ironman is slim) and the geometry from the previous stage. Our MV-ControlNet addresses this misalignment by understanding the depth information of the geometry, thereby enhancing alignment and coherence. 
While Fantasia3D and MVDream successfully stylize the overall geometry, they tend to produce results with issues like oversaturation or residual traits from the original geometry prompt. We invite the readers to explore additional results on the project page.

\subsection{Quantitative Comparison of 3D Generation}
\subsubsection{Directional CLIP Similarity}
To quantitatively compare the alignment between prompt modifications and image changes in ControlDreamer and MVDream, we evaluate the directional CLIP similarity \cite{gal2022stylegan} between first and second stage 3D models on our benchmark dataset. This score measures how well the change of rendered images align with the text directions within the CLIP space. We use texts with three view-oriented postfixes: `a front view', `a side view', and `a back view', utilizing four views from each azimuth angles (0°, 90°, 180°, 270°) for the 3D samples.

In Table~\ref{tab:dir_clip_mean}, averaged similarities and standard deviations are presented for each domain, across all six prompt candidates and directions. We employ the OpenCLIP ViT-L/14 model~\cite{cherti2023reproducible}, trained on the LAION-2B dataset, which is different from the text encoder used in ControlDreamer. Our method demonstrates superior performance, excelling in generating text-aligned 3D models across all evaluated domains in our benchmark. Detailed explanations of the used prompts are presented in the Appendix. %  \ref{tab:used_bench}

%%%%%%%%%%%
\begin{table}[t!]
\centering
\caption{Comparison of directional CLIP similarity on five different categories.}
\footnotesize
\label{tab:dir_clip_mean}
{\resizebox{0.6\textwidth}{!}
{\begin{tabular}{cccccc}
\hline
    \toprule
    Domain & Animals & Characters & Foods & General & Objects \\
    \midrule
    Magic3D & 0.218 & 0.280 & 0.236 & 0.247 & 0.217 \\
    Fantasia3D & 0.191 & 0.250  & 0.256 & 0.240 & 0.208  \\
    ProlificDreamer & 0.224 & 0.265  & 0.239 & 0.231 & 0.163 \\
    MVDream 2-stage & 0.160 & 0.235  & 0.150 & 0.211 & 0.122 \\
    \midrule
    ControlDreamer & \textbf{0.241} & \textbf{0.353}  & \textbf{0.270} & \textbf{0.341} & \textbf{0.255} \\
    \bottomrule
\end{tabular}}
}
\vspace{-1em}
\end{table} 

%%%%%

\subsubsection{Human Evaluations}
We assess preference scores using Amazon Mechanical Turk (AMT) across five domains in Fig. \ref{fig:human_eval}. As a result, the proposed method consistently achieves the highest preference scores in all domains, demonstrating its effectiveness. This assessment allows for a comparative ranking of MVDream, Fantasia3D, Magic3D, and ProlificDreamer. We assess each methodology using 30 3D models from each domain, with 200 human evaluators participating. Each evaluator encounters three random samples and repeats the evaluation task three times, resulting in about 20 evaluations per domain. For additional details, the templates for the human evaluation are included in the Appendix.
%\textcolor{blue}{We assessed text alignment based on direction using three post-fixes: `, a front view', `, a side view', and `, a back view'. Four different images of each viewing direction of the camera (\textit{i.e.}, 0-90-180-270 azimuth) from generated 3D models are used for evaluations.}
%%%%% Human Evaluation %%%%%%%%%%%%%%%%%%
%%%%%%%%%% arxiving 이후
% Human evaluation을 통해서 제안하는 방법이 다른 비교 방법들보다 더 월등하고 좋은 성능을 달성함을 보였다... (expected) -> evaluation 시 카테고리? 
% 6) Prompt-to-prompt를 3D에서 적용할 경우
% expected results) 하지만, pre-trained 3D controlnet (consistency-aware)이 부재하기 때문에 여전히 Janus problem, semantic drift같은 현상이 여전히 존재한다. 또한, mvdream baseline같은 경우는 texture를 잘 입히지 못하는 현상 발생. 
%%%%%%%%%%
\subsection{Ablation Study}
\subsubsection{Variants of MV-ControlNet}
In Fig. \ref{fig:compare} (b), we compare the 3D stylization results using MV-ControlNet variants trained under canny edge, normal, and depth conditions. With canny edge inputs, significant artifacts appear in both RGB and normals. When using normal maps, the RGB colors in the 3D models are less photorealistic. However, depth maps consistently produce 3D models without artifacts in both color and normal maps, showcasing their superiority.
%%%%%%%%%%%%%%%
\subsubsection{MV-ControlNet as a plug-in for prior methods}
Additionally, we demonstrate a scenario where MV-ControlNet is used in the refinement stage of existing methodologies, with identical geometry and style prompts. DreamFusion \cite{poole2022dreamfusion} is employed to create the geometry model. We then compare the performance of MV-ControlNet against Magic3D \cite{lin2023magic3d}. Fig. \ref{fig:compare} (d) highlights MV-ControlNet's adaptability as a plug-in component across various methods.
%%%%%%%%%%
%% 아카이빙 이후!
% \subsection{Human Evaluations}
% Pi chart로 제안하는 방법이 Animals, Characters, Foods, Objects 각각에서 baseline보다 월등히 성능이 높음을 확인하였다. 
% - baseline : magic3D, prolificdreamer, fantasia3d \\
% -photorealism,  quality 세가지 정도 생각중. 
% (3D) DALL-E-3 template으로 해당 데이터에 대한 LLM-based preference, correctness 실험 \\
% 2) ChatGPT한테 preference 정도 물어보고 판단 \\
% 3) 3D consistency에 대한 human evalution : 각 카테고리에 대한 sample들 뽑아서 AMT를 통해 preferene를 확인하였다. 

\section{Discussion}
We introduce ControlDreamer, a two-stage pipeline to improve text alignment in text-to-3D by decomposing the text into geometry and style prompts. To blend the geometry and style, we train MV-ControlNet, a depth-aware multi-view diffusion model, to enhance the geometry understanding during score distillation in the second stage. This integration effectively generates stylized 3D models by aligning diverse geometries and styles given by textual descriptions. Comparative results and theoretical analysis confirm that ControlDreamer outperforms existing text-to-3D methods in stylizing 3D models. Additionally, through comparisons with normal map and canny edge variants of MV-ControlNet, we demonstrated that depth conditioning is the most suitable for blending geometry and style.
%Additionally, our extensive benchmark for 3D style editing, covering a broad range of subjects, establishes a new standard in the field. 

While we improved text alignment in text-to-3D generation, some limitations remain. First, our training on single-object datasets limited our ability to experiment with complex scenes typical in 3D reconstruction. If multi-view diffusion models like ReconFusion~\cite{wu2024reconfusion}, trained on complex scenes, become publicly available, we could better assess ControlDreamer’s effectiveness. Additionally, though not our primary focus, future research could explore on various NeRF resolutions as these multi-view diffusion models advance to multiple resolutions and aspect ratios, similar to recent 2D image generative models~\cite{podell2023sdxl}. \\
%\newline
\clearpage
\newpage

\bibliography{main}

\begin{thebibliography}{44}
\providecommand{\natexlab}[1]{#1}
\providecommand{\url}[1]{\texttt{#1}}
\expandafter\ifx\csname urlstyle\endcsname\relax
  \providecommand{\doi}[1]{doi: #1}\else
  \providecommand{\doi}{doi: \begingroup \urlstyle{rm}\Url}\fi

\bibitem[Balaji et~al.(2022)Balaji, Nah, Huang, Vahdat, Song, Kreis, Aittala, Aila, Laine, Catanzaro, et~al.]{balaji2022ediffi}
Yogesh Balaji, Seungjun Nah, Xun Huang, Arash Vahdat, Jiaming Song, Karsten Kreis, Miika Aittala, Timo Aila, Samuli Laine, Bryan Catanzaro, et~al.
\newblock ediffi: Text-to-image diffusion models with an ensemble of expert denoisers.
\newblock \emph{arXiv preprint arXiv:2211.01324}, 2022.

\bibitem[Boss et~al.(2024)Boss, Huang, Vasishta, and Jampani]{boss2024sf3d}
Mark Boss, Zixuan Huang, Aaryaman Vasishta, and Varun Jampani.
\newblock Sf3d: Stable fast 3d mesh reconstruction with uv-unwrapping and illumination disentanglement.
\newblock \emph{arXiv preprint arXiv:2408.00653}, 2024.

\bibitem[Chen et~al.(2024{\natexlab{a}})Chen, Yang, Yang, Feng, Fu, Foo, Lin, and Liu]{chen2024sculpt3d}
Cheng Chen, Xiaofeng Yang, Fan Yang, Chengzeng Feng, Zhoujie Fu, Chuan-Sheng Foo, Guosheng Lin, and Fayao Liu.
\newblock Sculpt3d: Multi-view consistent text-to-3d generation with sparse 3d prior.
\newblock In \emph{Proceedings of the IEEE/CVF Conference on Computer Vision and Pattern Recognition}, pages 10228--10237, 2024{\natexlab{a}}.

\bibitem[Chen et~al.(2023{\natexlab{a}})Chen, Siddiqui, Lee, Tulyakov, and Nie{\ss}ner]{chen2023text2tex}
Dave~Zhenyu Chen, Yawar Siddiqui, Hsin-Ying Lee, Sergey Tulyakov, and Matthias Nie{\ss}ner.
\newblock Text2tex: Text-driven texture synthesis via diffusion models.
\newblock In \emph{Proceedings of the IEEE/CVF International Conference on Computer Vision (ICCV)}, pages 18558--18568, October 2023{\natexlab{a}}.

\bibitem[Chen et~al.(2023{\natexlab{b}})Chen, Chen, Jiao, and Jia]{chen2023fantasia3d}
Rui Chen, Yongwei Chen, Ningxin Jiao, and Kui Jia.
\newblock Fantasia3d: Disentangling geometry and appearance for high-quality text-to-3d content creation.
\newblock In \emph{Proceedings of the IEEE/CVF International Conference on Computer Vision (ICCV)}, pages 22246--22256, October 2023{\natexlab{b}}.

\bibitem[Chen et~al.(2024{\natexlab{b}})Chen, Zhang, Yang, Cai, Yu, Yang, and Lin]{chen2024it3d}
Yiwen Chen, Chi Zhang, Xiaofeng Yang, Zhongang Cai, Gang Yu, Lei Yang, and Guosheng Lin.
\newblock It3d: Improved text-to-3d generation with explicit view synthesis.
\newblock In \emph{Proceedings of the AAAI Conference on Artificial Intelligence}, volume~38, pages 1237--1244, 2024{\natexlab{b}}.

\bibitem[Cherti et~al.(2023)Cherti, Beaumont, Wightman, Wortsman, Ilharco, Gordon, Schuhmann, Schmidt, and Jitsev]{cherti2023reproducible}
Mehdi Cherti, Romain Beaumont, Ross Wightman, Mitchell Wortsman, Gabriel Ilharco, Cade Gordon, Christoph Schuhmann, Ludwig Schmidt, and Jenia Jitsev.
\newblock Reproducible scaling laws for contrastive language-image learning.
\newblock In \emph{Proceedings of the IEEE/CVF Conference on Computer Vision and Pattern Recognition}, pages 2818--2829, 2023.

\bibitem[Deitke et~al.(2023)Deitke, Schwenk, Salvador, Weihs, Michel, VanderBilt, Schmidt, Ehsani, Kembhavi, and Farhadi]{deitke2023objaverse}
Matt Deitke, Dustin Schwenk, Jordi Salvador, Luca Weihs, Oscar Michel, Eli VanderBilt, Ludwig Schmidt, Kiana Ehsani, Aniruddha Kembhavi, and Ali Farhadi.
\newblock Objaverse: A universe of annotated 3d objects.
\newblock In \emph{Proceedings of the IEEE/CVF Conference on Computer Vision and Pattern Recognition}, pages 13142--13153, 2023.

\bibitem[Gal et~al.(2022)Gal, Patashnik, Maron, Bermano, Chechik, and Cohen-Or]{gal2022stylegan}
Rinon Gal, Or~Patashnik, Haggai Maron, Amit~H Bermano, Gal Chechik, and Daniel Cohen-Or.
\newblock Stylegan-nada: Clip-guided domain adaptation of image generators.
\newblock \emph{ACM Transactions on Graphics (TOG)}, 41\penalty0 (4):\penalty0 1--13, 2022.

\bibitem[Guo et~al.(2023)Guo, Liu, Shao, Laforte, Voleti, Luo, Chen, Zou, Wang, Cao, and Zhang]{threestudio2023}
Yuan-Chen Guo, Ying-Tian Liu, Ruizhi Shao, Christian Laforte, Vikram Voleti, Guan Luo, Chia-Hao Chen, Zi-Xin Zou, Chen Wang, Yan-Pei Cao, and Song-Hai Zhang.
\newblock threestudio: A unified framework for 3d content generation.
\newblock \url{https://github.com/threestudio-project/threestudio}, 2023.

\bibitem[Haque et~al.(2023)Haque, Tancik, Efros, Holynski, and Kanazawa]{haque2023instruct}
Ayaan Haque, Matthew Tancik, Alexei~A. Efros, Aleksander Holynski, and Angjoo Kanazawa.
\newblock Instruct-nerf2nerf: Editing 3d scenes with instructions.
\newblock In \emph{Proceedings of the IEEE/CVF International Conference on Computer Vision (ICCV)}, pages 19740--19750, October 2023.

\bibitem[Hertz et~al.(2023)Hertz, Mokady, Tenenbaum, Aberman, Pritch, and Cohen-Or]{hertz2022prompt}
Amir Hertz, Ron Mokady, Jay Tenenbaum, Kfir Aberman, Yael Pritch, and Daniel Cohen-Or.
\newblock Prompt-to-prompt image editing with cross attention control.
\newblock In \emph{International Conference on Learning Representations}, 2023.

\bibitem[Ho and Salimans(2021)]{ho2022classifier}
Jonathan Ho and Tim Salimans.
\newblock Classifier-free diffusion guidance.
\newblock \emph{NeurIPS 2021 Workshop on Deep Generative Models and Downstream Applications}, 2021.

\bibitem[Ho et~al.(2020)Ho, Jain, and Abbeel]{ho2020denoising}
Jonathan Ho, Ajay Jain, and Pieter Abbeel.
\newblock Denoising diffusion probabilistic models.
\newblock \emph{Advances in neural information processing systems}, 33:\penalty0 6840--6851, 2020.

\bibitem[Katzir et~al.(2023)Katzir, Patashnik, Cohen-Or, and Lischinski]{katzir2023nfsd}
Oren Katzir, Or~Patashnik, Daniel Cohen-Or, and Dani Lischinski.
\newblock Noise-free score distillation.
\newblock \emph{arXiv preprint arXiv:2310.17590}, 2023.

\bibitem[Kingma et~al.(2021)Kingma, Salimans, Poole, and Ho]{kingma2021variational}
Diederik Kingma, Tim Salimans, Ben Poole, and Jonathan Ho.
\newblock Variational diffusion models.
\newblock \emph{Advances in neural information processing systems}, 34:\penalty0 21696--21707, 2021.

\bibitem[Kingma and Welling(2013)]{kingma2013auto}
Diederik~P Kingma and Max Welling.
\newblock Auto-encoding variational bayes.
\newblock \emph{arXiv preprint arXiv:1312.6114}, 2013.

\bibitem[Laine et~al.(2020)Laine, Hellsten, Karras, Seol, Lehtinen, and Aila]{Laine2020diffrast}
Samuli Laine, Janne Hellsten, Tero Karras, Yeongho Seol, Jaakko Lehtinen, and Timo Aila.
\newblock Modular primitives for high-performance differentiable rendering.
\newblock \emph{ACM Transactions on Graphics}, 39\penalty0 (6), 2020.

\bibitem[Li et~al.(2022)Li, Li, Xiong, and Hoi]{li2022blip}
Junnan Li, Dongxu Li, Caiming Xiong, and Steven Hoi.
\newblock Blip: Bootstrapping language-image pre-training for unified vision-language understanding and generation.
\newblock In \emph{International Conference on Machine Learning}, pages 12888--12900. PMLR, 2022.

\bibitem[Lin et~al.(2023)Lin, Gao, Tang, Takikawa, Zeng, Huang, Kreis, Fidler, Liu, and Lin]{lin2023magic3d}
Chen-Hsuan Lin, Jun Gao, Luming Tang, Towaki Takikawa, Xiaohui Zeng, Xun Huang, Karsten Kreis, Sanja Fidler, Ming-Yu Liu, and Tsung-Yi Lin.
\newblock Magic3d: High-resolution text-to-3d content creation.
\newblock In \emph{Proceedings of the IEEE/CVF Conference on Computer Vision and Pattern Recognition}, pages 300--309, 2023.

\bibitem[Liu et~al.(2023)Liu, Shi, Kuang, Zhu, Li, Han, Cai, Porikli, and Su]{liu2023openshape}
Minghua Liu, Ruoxi Shi, Kaiming Kuang, Yinhao Zhu, Xuanlin Li, Shizhong Han, Hong Cai, Fatih Porikli, and Hao Su.
\newblock Openshape: Scaling up 3d shape representation towards open-world understanding.
\newblock \emph{Advances in Neural Information Processing Systems}, 36, 2023.

\bibitem[Metzer et~al.(2023)Metzer, Richardson, Patashnik, Giryes, and Cohen-Or]{metzer2023latent}
Gal Metzer, Elad Richardson, Or~Patashnik, Raja Giryes, and Daniel Cohen-Or.
\newblock Latent-nerf for shape-guided generation of 3d shapes and textures.
\newblock In \emph{Proceedings of the IEEE/CVF Conference on Computer Vision and Pattern Recognition}, pages 12663--12673, 2023.

\bibitem[Mildenhall et~al.(2021)Mildenhall, Srinivasan, Tancik, Barron, Ramamoorthi, and Ng]{mildenhall2021nerf}
Ben Mildenhall, Pratul~P Srinivasan, Matthew Tancik, Jonathan~T Barron, Ravi Ramamoorthi, and Ren Ng.
\newblock Nerf: Representing scenes as neural radiance fields for view synthesis.
\newblock \emph{Communications of the ACM}, 65\penalty0 (1):\penalty0 99--106, 2021.

\bibitem[M\"uller et~al.(2022)M\"uller, Evans, Schied, and Keller]{mueller2022instant}
Thomas M\"uller, Alex Evans, Christoph Schied, and Alexander Keller.
\newblock Instant neural graphics primitives with a multiresolution hash encoding.
\newblock \emph{ACM Trans. Graph.}, 41\penalty0 (4):\penalty0 102:1--102:15, July 2022.
\newblock \doi{10.1145/3528223.3530127}.
\newblock URL \url{https://doi.org/10.1145/3528223.3530127}.

\bibitem[Munkberg et~al.(2022)Munkberg, Hasselgren, Shen, Gao, Chen, Evans, M\"uller, and Fidler]{Munkberg2022nvdiffrec}
Jacob Munkberg, Jon Hasselgren, Tianchang Shen, Jun Gao, Wenzheng Chen, Alex Evans, Thomas M\"uller, and Sanja Fidler.
\newblock {Extracting Triangular 3D Models, Materials, and Lighting From Images}.
\newblock In \emph{Proceedings of the IEEE/CVF Conference on Computer Vision and Pattern Recognition (CVPR)}, pages 8280--8290, June 2022.

\bibitem[Oh et~al.(2024)Oh, Lee, Choi, Jung, Hwang, and Yoon]{oh2024efficient}
Yeongtak Oh, Jonghyun Lee, Jooyoung Choi, Dahuin Jung, Uiwon Hwang, and Sungroh Yoon.
\newblock Efficient diffusion-driven corruption editor for test-time adaptation.
\newblock \emph{arXiv preprint arXiv:2403.10911}, 2024.

\bibitem[Oord et~al.(2018)Oord, Li, Babuschkin, Simonyan, Vinyals, Kavukcuoglu, Driessche, Lockhart, Cobo, Stimberg, et~al.]{oord2018parallel}
Aaron Oord, Yazhe Li, Igor Babuschkin, Karen Simonyan, Oriol Vinyals, Koray Kavukcuoglu, George Driessche, Edward Lockhart, Luis Cobo, Florian Stimberg, et~al.
\newblock Parallel wavenet: Fast high-fidelity speech synthesis.
\newblock In \emph{International conference on machine learning}, pages 3918--3926. PMLR, 2018.

\bibitem[OpenAI(2023)]{gpt4}
OpenAI.
\newblock Gpt-4 technical report.
\newblock 2023.

\bibitem[Podell et~al.(2023)Podell, English, Lacey, Blattmann, Dockhorn, M{\"u}ller, Penna, and Rombach]{podell2023sdxl}
Dustin Podell, Zion English, Kyle Lacey, Andreas Blattmann, Tim Dockhorn, Jonas M{\"u}ller, Joe Penna, and Robin Rombach.
\newblock Sdxl: Improving latent diffusion models for high-resolution image synthesis.
\newblock \emph{arXiv preprint arXiv:2307.01952}, 2023.

\bibitem[Poole et~al.(2023)Poole, Jain, Barron, and Mildenhall]{poole2022dreamfusion}
Ben Poole, Ajay Jain, Jonathan~T Barron, and Ben Mildenhall.
\newblock Dreamfusion: Text-to-3d using 2d diffusion.
\newblock In \emph{International Conference on Learning Representations}, 2023.

\bibitem[Radford et~al.(2021)Radford, Kim, Hallacy, Ramesh, Goh, Agarwal, Sastry, Askell, Mishkin, Clark, et~al.]{radford2021learning}
Alec Radford, Jong~Wook Kim, Chris Hallacy, Aditya Ramesh, Gabriel Goh, Sandhini Agarwal, Girish Sastry, Amanda Askell, Pamela Mishkin, Jack Clark, et~al.
\newblock Learning transferable visual models from natural language supervision.
\newblock In \emph{International conference on machine learning}, pages 8748--8763. PMLR, 2021.

\bibitem[Ranftl et~al.(2022)Ranftl, Lasinger, Hafner, Schindler, and Koltun]{Ranftl2022midas}
Ren\'{e} Ranftl, Katrin Lasinger, David Hafner, Konrad Schindler, and Vladlen Koltun.
\newblock Towards robust monocular depth estimation: Mixing datasets for zero-shot cross-dataset transfer.
\newblock \emph{IEEE Transactions on Pattern Analysis and Machine Intelligence}, 44\penalty0 (3), 2022.

\bibitem[Richardson et~al.(2023)Richardson, Metzer, Alaluf, Giryes, and Cohen-Or]{richardson2023texture}
Elad Richardson, Gal Metzer, Yuval Alaluf, Raja Giryes, and Daniel Cohen-Or.
\newblock Texture: Text-guided texturing of 3d shapes.
\newblock \emph{arXiv preprint arXiv:2302.01721}, 2023.

\bibitem[Rombach et~al.(2022)Rombach, Blattmann, Lorenz, Esser, and Ommer]{rombach2022high}
Robin Rombach, Andreas Blattmann, Dominik Lorenz, Patrick Esser, and Bj{\"o}rn Ommer.
\newblock High-resolution image synthesis with latent diffusion models.
\newblock In \emph{Proceedings of the IEEE/CVF conference on computer vision and pattern recognition}, pages 10684--10695, 2022.

\bibitem[Ronneberger et~al.(2015)Ronneberger, Fischer, and Brox]{ronneberger2015unet}
Olaf Ronneberger, Philipp Fischer, and Thomas Brox.
\newblock U-net: Convolutional networks for biomedical image segmentation.
\newblock In \emph{Medical Image Computing and Computer-Assisted Intervention--MICCAI 2015: 18th International Conference, Munich, Germany, October 5-9, 2015, Proceedings, Part III 18}, pages 234--241. Springer, 2015.

\bibitem[Saharia et~al.(2022)Saharia, Chan, Saxena, Li, Whang, Denton, Ghasemipour, Gontijo~Lopes, Karagol~Ayan, Salimans, et~al.]{saharia2022photorealistic}
Chitwan Saharia, William Chan, Saurabh Saxena, Lala Li, Jay Whang, Emily~L Denton, Kamyar Ghasemipour, Raphael Gontijo~Lopes, Burcu Karagol~Ayan, Tim Salimans, et~al.
\newblock Photorealistic text-to-image diffusion models with deep language understanding.
\newblock \emph{Advances in Neural Information Processing Systems}, 35:\penalty0 36479--36494, 2022.

\bibitem[Schuhmann et~al.(2022)Schuhmann, Beaumont, Vencu, Gordon, Wightman, Cherti, Coombes, Katta, Mullis, Wortsman, et~al.]{schuhmann2022laion}
Christoph Schuhmann, Romain Beaumont, Richard Vencu, Cade Gordon, Ross Wightman, Mehdi Cherti, Theo Coombes, Aarush Katta, Clayton Mullis, Mitchell Wortsman, et~al.
\newblock Laion-5b: An open large-scale dataset for training next generation image-text models.
\newblock \emph{Advances in Neural Information Processing Systems}, 35:\penalty0 25278--25294, 2022.

\bibitem[Shen et~al.(2021)Shen, Gao, Yin, Liu, and Fidler]{shen2021deep}
Tianchang Shen, Jun Gao, Kangxue Yin, Ming-Yu Liu, and Sanja Fidler.
\newblock Deep marching tetrahedra: a hybrid representation for high-resolution 3d shape synthesis.
\newblock \emph{Advances in Neural Information Processing Systems}, 34:\penalty0 6087--6101, 2021.

\bibitem[Shi et~al.(2023)Shi, Wang, Ye, Long, Li, and Yang]{shi2023mvdream}
Yichun Shi, Peng Wang, Jianglong Ye, Mai Long, Kejie Li, and Xiao Yang.
\newblock Mvdream: Multi-view diffusion for 3d generation.
\newblock \emph{arXiv preprint arXiv:2308.16512}, 2023.

\bibitem[Voleti et~al.(2024)Voleti, Yao, Boss, Letts, Pankratz, Tochilkin, Laforte, Rombach, and Jampani]{voleti2024sv3d}
Vikram Voleti, Chun-Han Yao, Mark Boss, Adam Letts, David Pankratz, Dmitry Tochilkin, Christian Laforte, Robin Rombach, and Varun Jampani.
\newblock Sv3d: Novel multi-view synthesis and 3d generation from a single image using latent video diffusion.
\newblock \emph{arXiv preprint arXiv:2403.12008}, 2024.

\bibitem[Wang and Shi(2023)]{wang2023imagedream}
Peng Wang and Yichun Shi.
\newblock Imagedream: Image-prompt multi-view diffusion for 3d generation.
\newblock \emph{arXiv preprint arXiv:2312.02201}, 2023.

\bibitem[Wang et~al.(2023)Wang, Lu, Wang, Bao, Li, Su, and Zhu]{wang2023prolificdreamer}
Zhengyi Wang, Cheng Lu, Yikai Wang, Fan Bao, Chongxuan Li, Hang Su, and Jun Zhu.
\newblock Prolificdreamer: High-fidelity and diverse text-to-3d generation with variational score distillation.
\newblock \emph{Advances in Neural Information Processing Systems}, 36, 2023.

\bibitem[Wu et~al.(2024)Wu, Mildenhall, Henzler, Park, Gao, Watson, Srinivasan, Verbin, Barron, Poole, et~al.]{wu2024reconfusion}
Rundi Wu, Ben Mildenhall, Philipp Henzler, Keunhong Park, Ruiqi Gao, Daniel Watson, Pratul~P Srinivasan, Dor Verbin, Jonathan~T Barron, Ben Poole, et~al.
\newblock Reconfusion: 3d reconstruction with diffusion priors.
\newblock In \emph{Proceedings of the IEEE/CVF Conference on Computer Vision and Pattern Recognition}, pages 21551--21561, 2024.

\bibitem[Zhang et~al.(2023)Zhang, Rao, and Agrawala]{zhang2023adding}
Lvmin Zhang, Anyi Rao, and Maneesh Agrawala.
\newblock Adding conditional control to text-to-image diffusion models.
\newblock In \emph{Proceedings of the IEEE/CVF International Conference on Computer Vision}, pages 3836--3847, 2023.

\end{thebibliography}

\section*{Acknowledgement}
\noindent This research was supported by the Institute of Information \& Communications Technology Planning \& Evaluation (IITP) through a grant from the Korea government (MSIT) [No. RS-2021-II211343, Artificial Intelligence Graduate School Program (Seoul National University)], the National Research Foundation of Korea (NRF) grant from the Korea government (MSIT) (No. 2022R1A3B1077720 and No. 2022R1A5A7083908), and the BK21 FOUR program for the Education and Research Program for Future ICT Pioneers at Seoul National University in 2024.

\appendix
% Source, target
% Corgi seed 0, 2023
% Astronaut seed 8888, 
%%%%
% \input{Tabs/tab3}

%%%%%%
% \subsubsection{Ablation on editing scheduling}
% \textcolor{red}{공정한 실험인지 의논 필요. "conventional method"는 아예 depth 컨디션이 없는데 depth condition 못 따른다고 비교해도 되는건지? 차라리 depth 컨디션 쓰면 counterfactual 샘플 생성 가능하다고 보여주는건 어떤지? (지금의 Fig.4 (B)처럼) 그리고 Figure 4도 재배치 필요.}
% arxiving 이후 생각

%%%%%%%%%
\section{Additional Experiments}
\subsection{Qualitative results} 
\begin{figure*}[ht!]
\centering
\includegraphics[width=\linewidth]{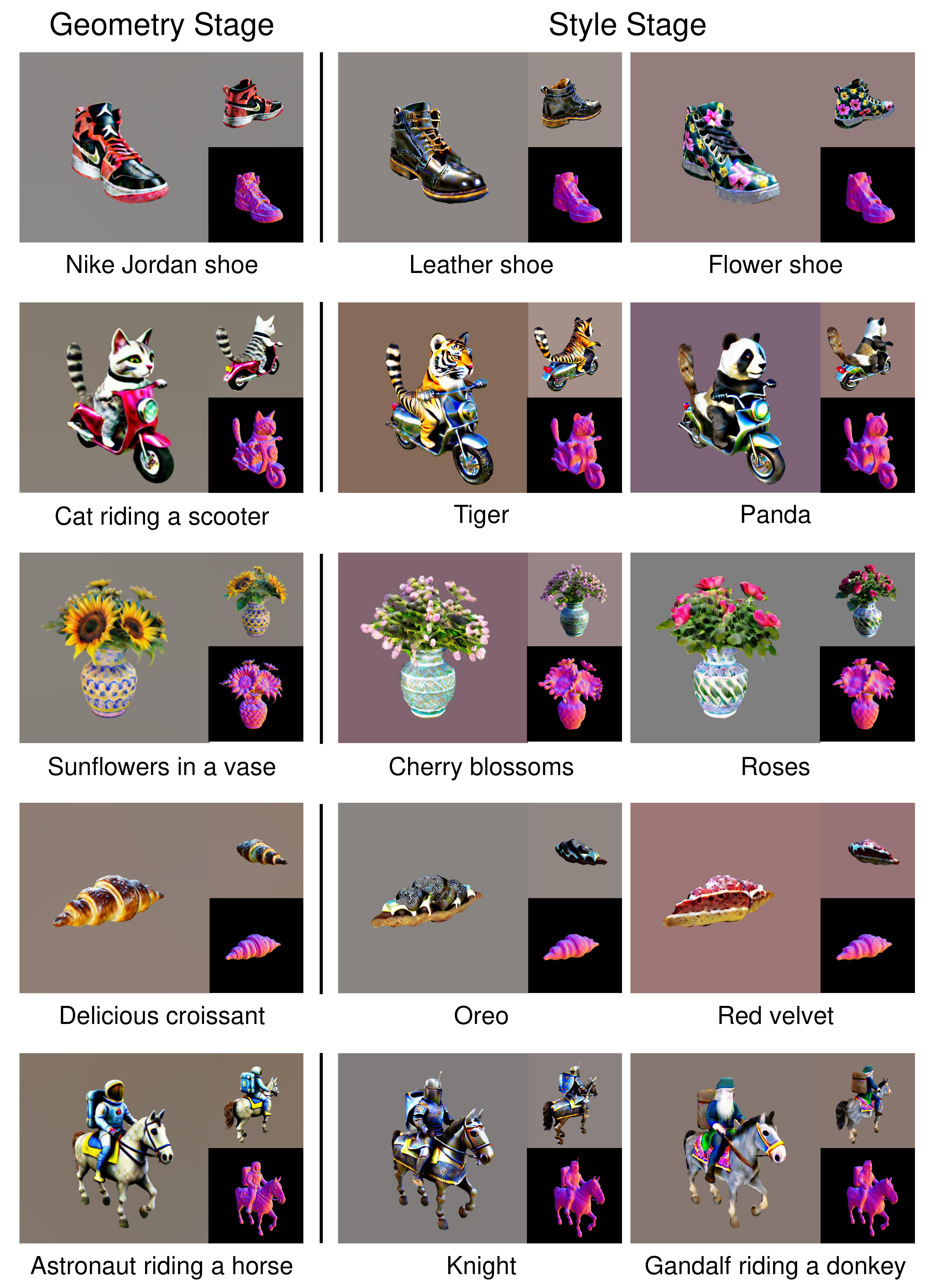} 
\caption{Our two-stage 3D generation pipeline first involves generating a coarse-grained geometry, followed by the generation of a fine-grained stylized 3D model using a style prompt.}
\label{fig:style_mixing}
\end{figure*}

\begin{figure*}[t!]
\centering
\includegraphics[width=\linewidth]{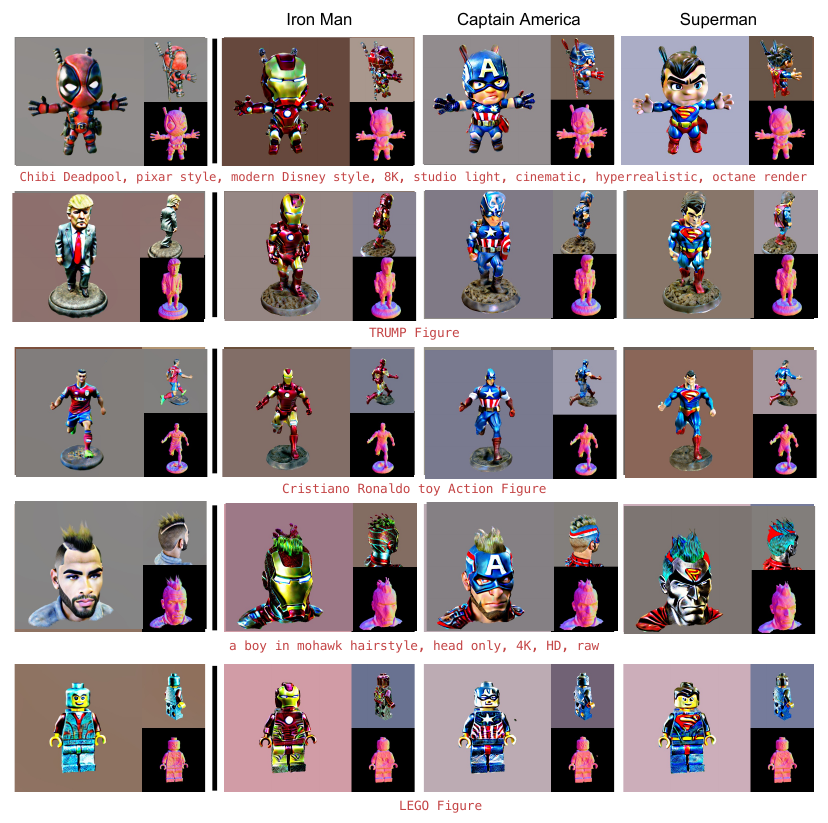} 
\caption{Additional qualitative results. We generate a variety of styles on a range of source 3D models, demonstrating the versatility of our method.}
\label{fig:additional_style}
%%%
\end{figure*}

In Figs. \ref{fig:style_mixing} and \ref{fig:additional_style}, we present a variety of qualitative results used for evaluating directional CLIP similarities and conducting human evaluations. 
%%%%%%%%%%%%%%%%%%%%%
\subsection{Ablation Studies}
\subsubsection{Exploring 3D Representations in Style Stages}
\begin{figure*}[h!]
\includegraphics[width=\linewidth]{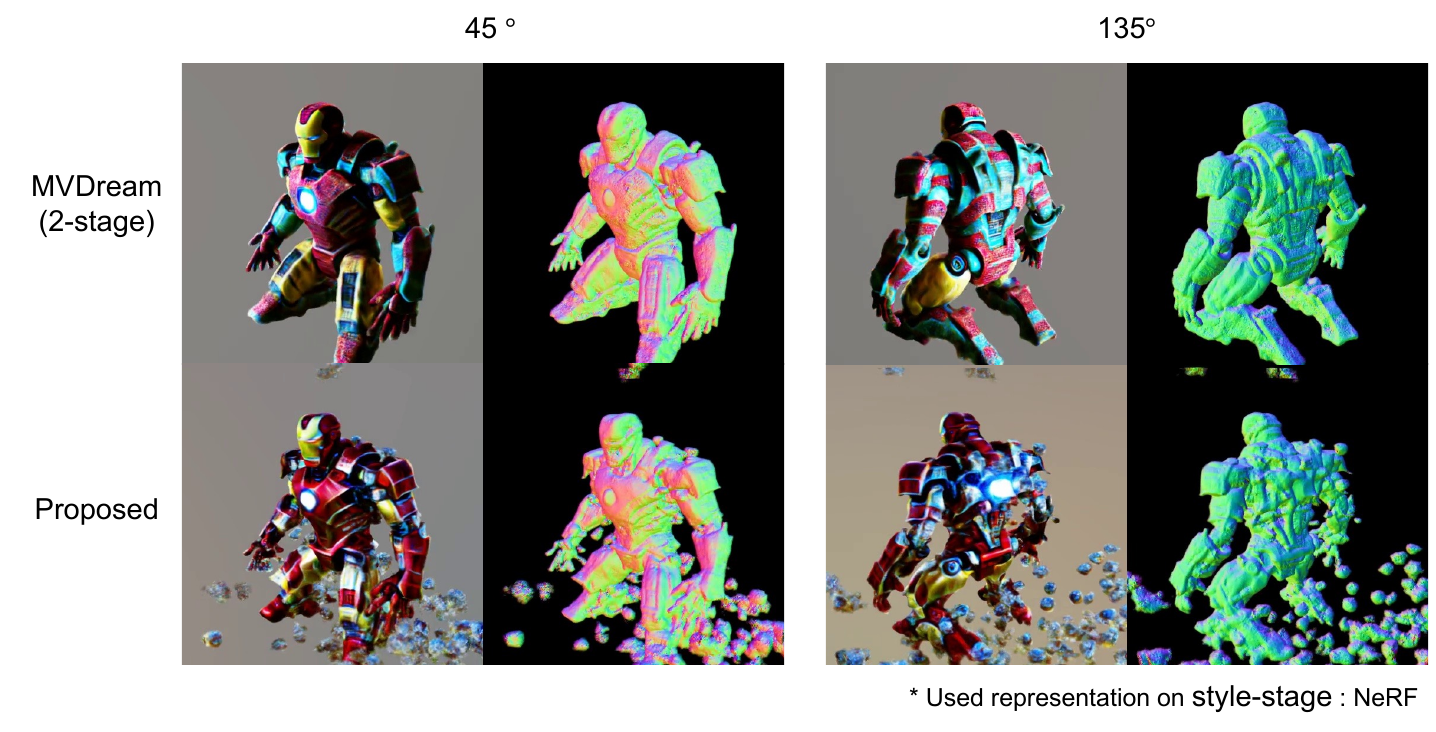} 
\caption{We visualize the results for both the two-stage MVDream and our ControlDreamer method in the style stage using NeRF representation. Notably, with NeRF as the representation, the geometry undergoes significant changes. This substantial alteration in geometry and depth information during NeRF training leads to instability in our depth-aware score distillation process, resulting in the generation of gravel-like artifacts in our results.}
\label{fig:nerf_abl}
\end{figure*}

Fig. \ref{fig:nerf_abl} demonstrates the approach of refining 3D models using NeRF instead of DMTet during the style stage. Volume rendering with NeRF often results in significant alterations to the overall geometry, particularly when geometry and style texts are semantically different, leading to instability in depth-aware score distillation. Conversely, using DMTet, coupled with our timestep annealing approach, is more effective. It ensures appropriate alterations in the overall texture while maintaining the geometry aligned with the source mesh, leading to more seamless style generation. For NeRF-based 3D stylization, we sample diffusion timesteps from a range of [0.02, 0.5] to minimize significant geometry changes. MV-ControlNet's tendency to overlook camera conditions often leads to multiple artifacts, further substantiating the advantage of using DMTet in the style stage.
%%%%%%%%%%%%%%%%%%%%
%%%%
%%%
\begin{figure*}[h!]
    \includegraphics[width=\linewidth]{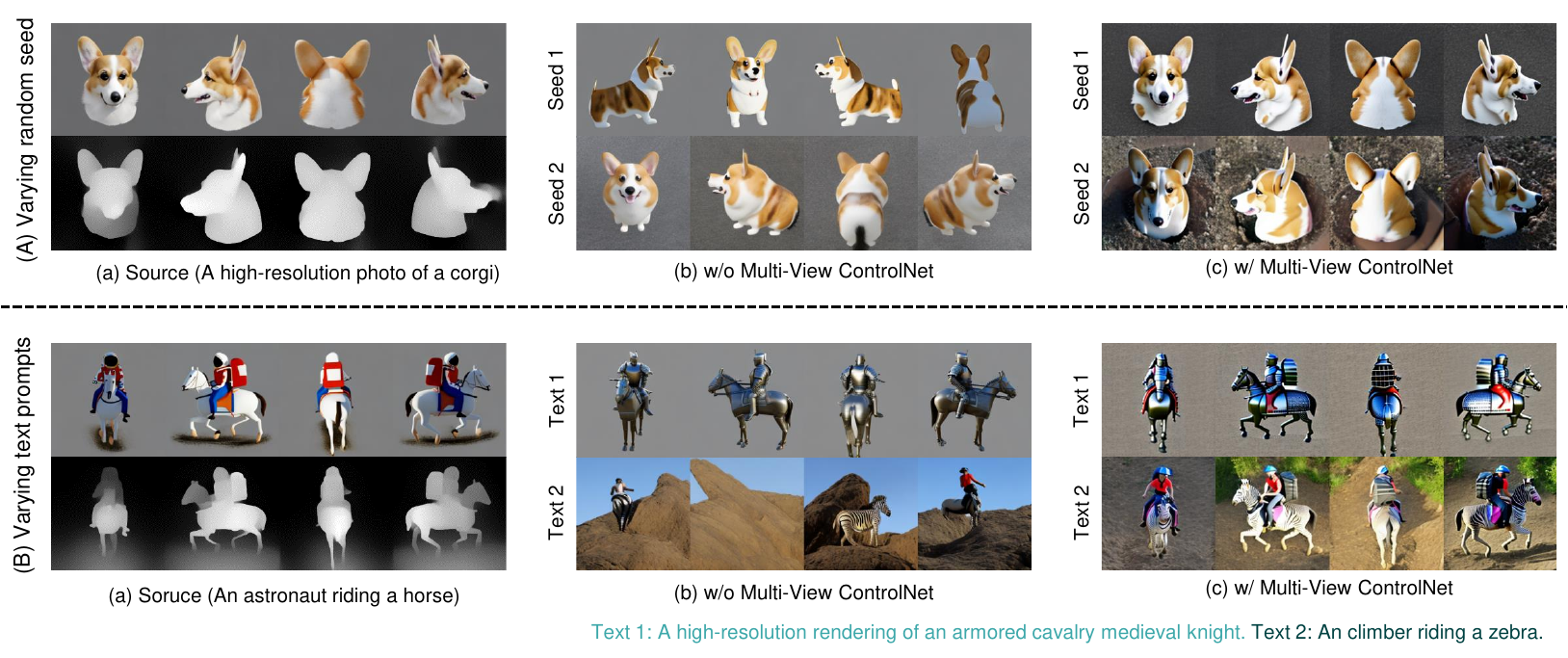} 
    \centering
    \caption{(A) The first column displays source multi-view images along with their corresponding depths. We present a comparison of images generated by MVDream in the second column and MV-ControlNet in the third column, originating from two distinct initial noises. MV-ControlNet's depth-aware generation consistently preserves the source's geometry and effectively manages variations in initial noise. (B) Our model excels at generating unique images, such as a knight equipped with a backpack, and adeptly handles imaginative prompts like `climber zebra'. It consistently produces high-quality images, outperforming MVDream, which often struggles with such complex tasks.}
    \label{fig:control_1}
\end{figure*}
%%%%%%%%%%%%%%%%%%%%%
\subsubsection{Advantages of MV-ControlNet in multi-view image generation}
Fig. \ref{fig:control_1} illustrates the advantages of the proposed MV-ControlNet in 2D image generation. In panel (A), conventional methods exhibit significant semantic variations due to random seed and camera parameter changes, often deviating from the original identity of the source image. In contrast, our proposed method demonstrates stronger alignment with the depth condition, ensuring consistent, object-centric generation quality across different camera parameters and random seeds. 

In panel (B), when handling text with complex concepts, the baseline method displays stereotypical and geometry-biased generations that compromise image quality, as discussed earlier in Fig. 1 of the main text. Conversely, our method consistently produces high-quality images that align closely with the provided text, preserving key identities from the source, such as the backpack of an astronaut.

In summary, MV-ControlNet is firmly grounded in depth conditions and adept at incorporating complex textual concepts into the generation of multi-view images.

\subsection{Effectiveness of Prompt Engineering for 3D Generation}
When using user-defined text prompts that were not included in the training set of the MVDream model, we observe the generation of artifacts in the learned source geometry, as illustrated in Fig.~\ref{fig:chatgpt}. However, we find that careful prompt engineering can effectively mitigate these artifacts. An example of such refined text in our experiments is `Hulk, muscular, green, 4K, photorealistic'.

%%%%
\subsection{Geometry bias in Objaverse dataset}
\begin{figure*}[t!]
\centering
\includegraphics[width=\linewidth]{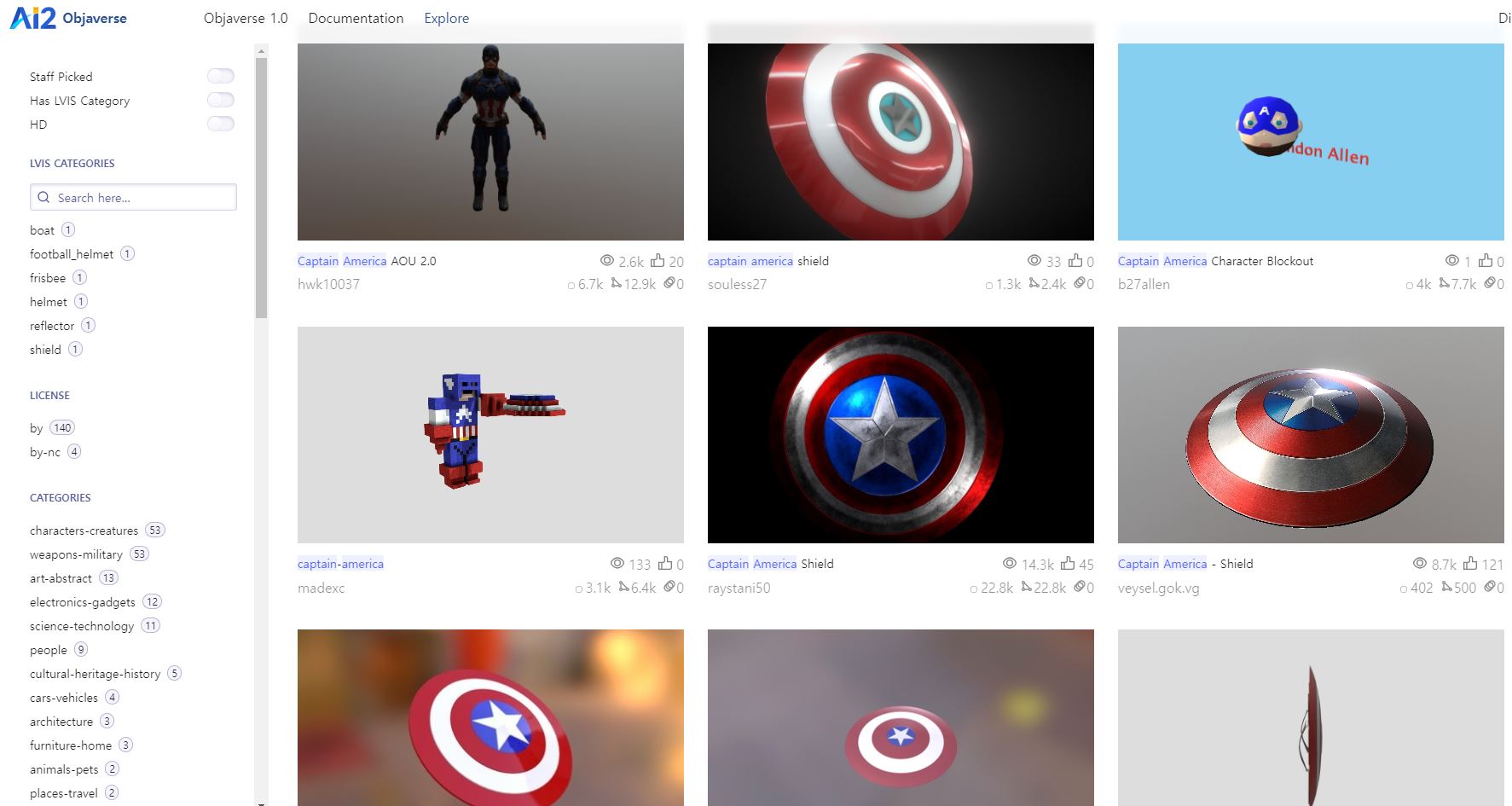} 
\caption{Intriguingly, a search for `Captain America' on \url{https://objaverse.allenai.org/explore} primarily yields content related to shields.}
\label{fig:objaverse}
\end{figure*}

\begin{figure*}[h!]
\includegraphics[width=\linewidth]{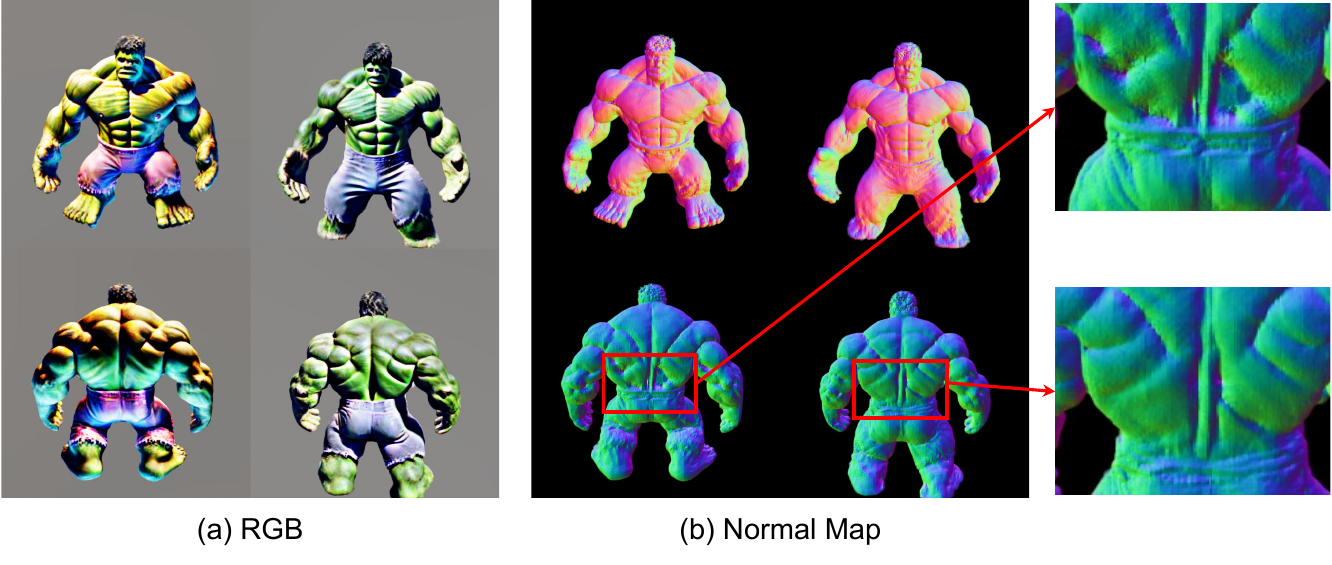} 
\caption{The comparison showcases 3D models generated from user-provided texts versus those recommended by ChatGPT. Notably, models from user texts may exhibit artifacts, particularly in normal maps, whereas models from ChatGPT's refined texts generally result in higher-quality 3D models.}
\label{fig:chatgpt}
\end{figure*}

We include further results on the geometry bias. Fig. \ref{fig:objaverse} provides a straightforward visualization of text-based 3D model retrieval on the Objaverse dataset, showcasing the frequency of certain geometries. Notably, as we highlight in the main text, 3D models predominantly associated with a shield are frequently generated from the prompt `Captain America'.

%%%%%%%
\subsection{Enhanced Stylization in the Image-to-3D Framework}
Our methodology demonstrates exceptional stylization capabilities, independent of the initialized geometry. Building on this strength, we have extended our experiments to include the image-to-3D framework as described in ImageDream~\cite{wang2023imagedream}, to further evaluate its editing capabilities. On 3D models generated using ImageDream, we applied our MV-ControlNet to modify the style. The results, depicted in Fig. \ref{fig:imagedream}, show the adaptability of our approach to the image-to-3D generation.
%%%%%%%%%%%%%%%%%%%%%%%%%%%%%%%%%%%%%%%%%%%%%
%%%%%%%%%%%%%%%%%%%%%%%%%%%%%%%%%%%%%%%%%%%%
%%%%
\begin{figure*}[t!]
    \centering
    \includegraphics[width=\linewidth]{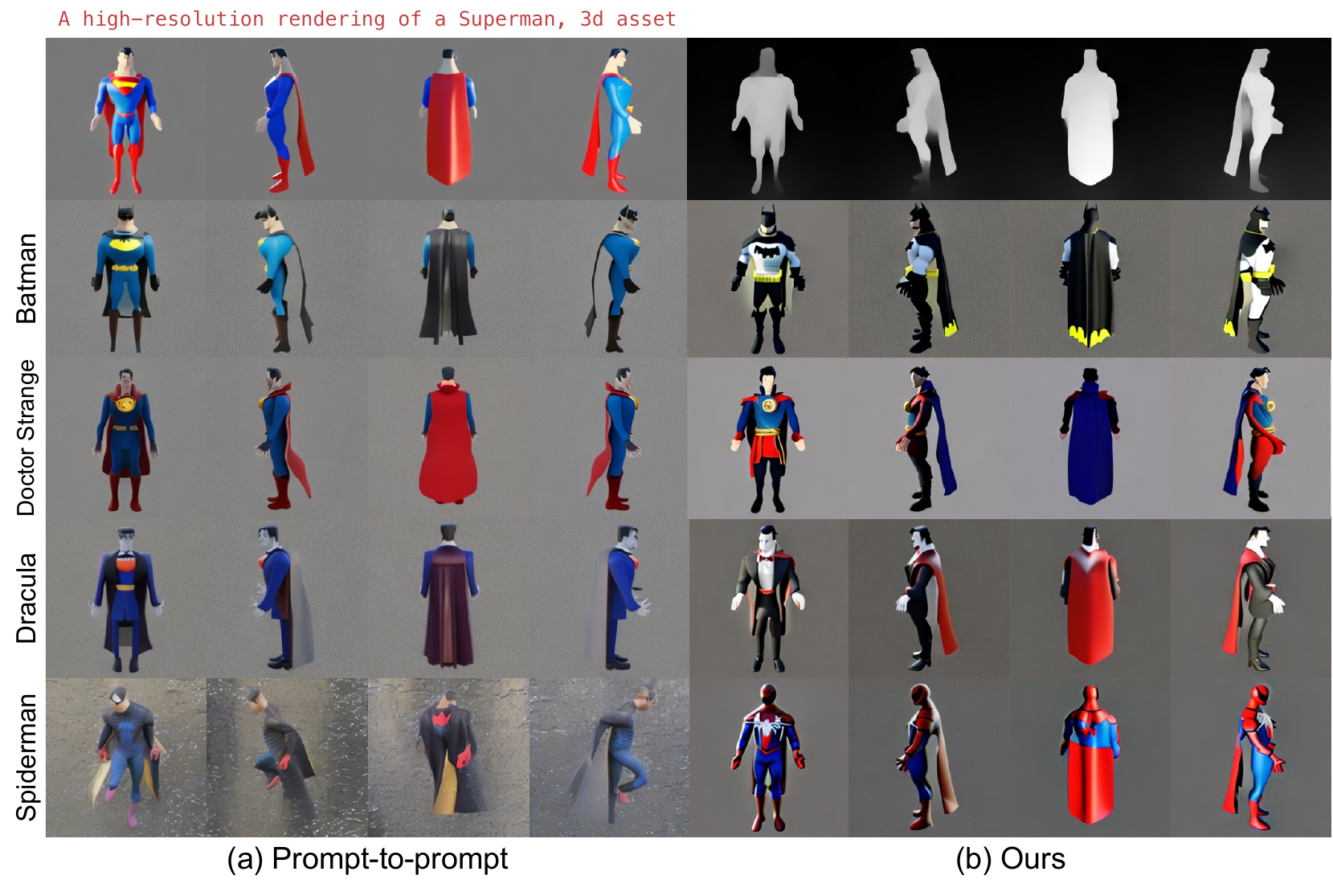} 
    \caption{The comparative results on multi-view image editings. Our method consistently generates more photorealistic results than P2P~\cite{hertz2022prompt}.}
    \label{fig:ex1}
\end{figure*}
%%%
\subsection{Multi-view image editing}
Prompt-to-Prompt (P2P) \cite{hertz2022prompt} is an image editing method that aligns the geometries of source and target images by injecting attention maps into diffusion models. For our qualitative comparison in the main text, injection ratios were set at 0.4 for cross-attention and 0.8 for self-attention to ensure strong geometric alignment. However, P2P is somewhat constrained by the requirement for equal token lengths in both source and target texts, which often results in less aligned text-guided editing outcomes. Conversely, our proposed MV-ControlNet, which utilizes depth conditions to maintain source geometry, is not restricted by text length, allowing for more creative and diverse prompts in the style stage.

In Fig. \ref{fig:ex1}, we showcase editing results of multi-view images using various prompts. Using `Superman' as a geometry prompt, we added a cloak to the initial geometry. The P2P method, which relies on cross-attention control, struggles with photorealistic images for complex scenarios like Spiderman wearing a cloak. Conversely, our MV-ControlNet incorporates depth conditions, achieving better style alignment and consistently producing high-quality, text-aligned images.
\subsection{Failure cases}
In Fig. \ref{fig:failure}, we present examples of our failure cases. We empirically observe that when the style prompt significantly deviates from the initial geometry, our method struggles to generate visually appealing 3D models. This limitation likely arises from the pre-trained MVDream, which has insufficient capabilities to seamlessly integrate two distinct concepts.

\section{Experiment Details}
\subsection{Baselines}
We utilize the threestudio \cite{threestudio2023} implementation of Magic3D \cite{lin2023magic3d}, Fantasia3D \cite{chen2023fantasia3d}, and ProlificDreamer \cite{wang2023prolificdreamer} as our baselines. Similar to ControlDreamer, all baselines involve a refinement process on textured meshes derived from the pre-trained MVDream \cite{shi2023mvdream} using the DMTet \cite{shen2021deep} algorithm. Specifically, Fantasia3D employs a two-stage process, focusing on appearance modeling in the second stage to modify the textured mesh. For ProlificDreamer, we engage in a textured mesh fine-tuning stage for stylization. Additionally, although ProlificDreamer does not utilize a separate shading model, we apply shading to ensure fair comparisons with other baselines. Our paired geometry and style prompts are detailed in Table \ref{tab:used_bench}.
%%%
\begin{table*}[t!]
\caption{Comprehensive descriptions of the utilized geometry and style prompts.}
\label{tab:used_bench}
{\resizebox{\textwidth}{!}
{\begin{tabular}{lll}
\hline
    \toprule   
    Domain & Geometry texts & Style texts \\
    \midrule
    Animals & A bald eagle carved out of wood & A highly-detailed rendering of a magpie, ultra hd, realistic, vivid colors \\
     & A high-resolution photo of a frog, 3d asset & A high-resolution photo of a toad, 3d asset\\
     & A cute shark, plush toy, ultra realistic, 4K, HD & A highly-detailed photo of a tuna, 4k, HD \\
     & A cute shark, plush toy, ultra realistic, 4K, HD & A highly-detailed rendering of a killer whale, 4k, HD \\
     & A high-resolution rendering of a Shiba Inu, 3d asset & A white tiger, super detailed, best quality, 4K, HD \\
     & A high-resolution photo of a British Shorthair, 3d asset & A highly-detailed photo of a lioness, 4K, HD\\
    \midrule    
    Character & A high-resolution rendering of a Hulk, 3d asset & A highly-detailed 3d rendering of a Superman \\
     & A high-resolution rendering of a Hulk, 3d asset & A high-resolution rendering of an Iron Man, 3d asset \\
     & A high-resolution rendering of a Hulk, 3d asset & A highly-detailed photo of a Spider-Man, 4K, HD \\
     & A high-resolution rendering of a Hulk, 3d asset & Renaissance sculpture of Michelangelo's David, Masterpiece \\
     & A high-resolution rendering of a Hulk, 3d asset & A highly-detailed 3d rendering of Thanos wearing the infinity gauntlet \\
     & A high-resolution rendering of a Hulk, 3d asset & A highly-detailed 3d rendering of a Captain America \\
    \midrule   
    Foods & A delicious croissant, realistic, 4K, HD & A delicious red velvet cake, realistic, 4K, HD \\
     & A delicious croissant, realistic, 4K, HD  & A delicious pecan pie, realistic, 4K, HD \\
     & A delicious croissant, realistic, 4K, HD  & A delicious oreo cake, realistic, 4K, HD \\
     & wedding cake, 4k & stack of macarons, 4k \\
     & wedding cake, 4k & strawberry cake, 4k \\
     & wedding cake, 4k & cheeseburger, 4k\\
    \midrule
    General & a cat riding a scooter like a human & panda riding a scooter like a human, 4k \\ 
     & a cat riding a scooter like a human  & tiger riding a scooter like a human, 4k \\
     & a cat riding a scooter like a human  & corgi riding a scooter like a human, 4k \\
     & An astronaut riding a horse & A high-resolution rendering of an armored cavalry medieval knight, 3d asset \\
     & An astronaut riding a horse & Gandalf riding a donkey, fairy tale style, 4K, HD \\
     & An astronaut riding a horse & A beautiful portrait of a princess riding an unicorn, fantasy, HD \\
    \midrule 
    Objects & A cute shark, plush toy, ultra realistic, 4K, HD & A DSLR photo of a submarine, 4k, HD \\
     & Battletech Zeus with a sword!, tabletop, miniature, & Optimus Prime fighting, super detailed, best quality, 4K, HD \\
     & battletech, miniatures, wargames, 3d asset & \\
     & Battletech Zeus with a sword!, tabletop, miniature, & A high-resolution rendering of a Gundam, 3d asset \\
     & battletech, miniatures, wargames, 3d asset & \\
     & Nike Jordan shoe, 4k & leather shoe, 4k \\
     & Nike Jordan shoe, 4k & shoe with flower printed on it, 4k \\
     & 3d rendering of a mug of hot chocolate with whipped cream & A highly-detailed photo of a mug with Starbucks logo, 4K, HD \\
    \bottomrule
\end{tabular}}}
\end{table*}
%%%
\subsection{Generated images used for MV-ControlNet training}
In Fig. \ref{fig:additional_1} and Fig. \ref{fig:additional_2}, we present additional examples used in training our MV-ControlNet. Both figures illustrate samples created through random sampling, with the camera views for each sample also randomized. These figures clearly show that samples from a filtered $100K$ text corpus feature a wide range of 3D assets, such as objects, animals, and characters. We used these high-quality multi-view images to train MV-ControlNet.
%%%%%%%%%
\subsection{Human evaluations}
Our human evaluation spans paired prompts from five domains, with each domain involving six prompts across five methodologies: Fantasia3D, Magic3D, ProlificDreamer, MVDream, and our approach. The evaluations assess editability and text alignment, as detailed in the table below.
\begin{mybox}{}
    Your task is to evaluate the quality of 4.0-second-long food video clips. Each clip showcases the results of editing 3D models using text prompts. \\
    Please choose the most favorable video based on the following criteria: \\
    
    \textcolor{purple}{1) Overall editing quantity}: Has the edited video completely changed in visual appearance compared to the Source? \\
    \textcolor{violet}{2) Textual alignment}: How well does the text description align with the edited video?
\end{mybox}

%%%%%% Figure 4 %%%%%%
%%%%%%%%%%%%
%%%%%%%%%%%%%%%%%%
\begin{figure*}[t!]
\centering
\includegraphics[width=\linewidth]{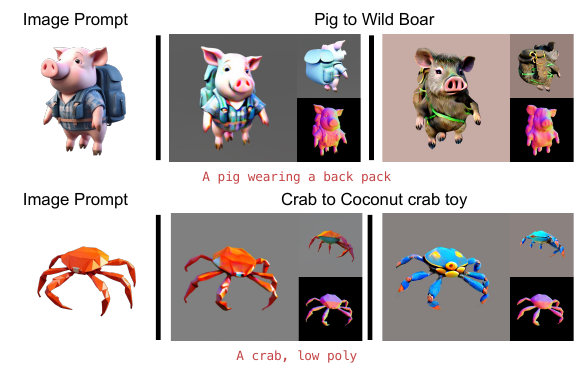} 
\caption{Visualization of our stylization process within the image-to-3D pipeline. The source image prompts chosen for this demonstration were the subjects `pig' and `crab'. Across these distinct subjects, our method maintained robust editing performance, showcasing its consistency in 3D stylization.}
\label{fig:imagedream}
\end{figure*}
%%%%%%%%%%%%%%%%%%%%%%%%%%%%%%%%%%%%%%%%%%%%
\begin{figure*}[h!]
    \includegraphics[width=\linewidth]{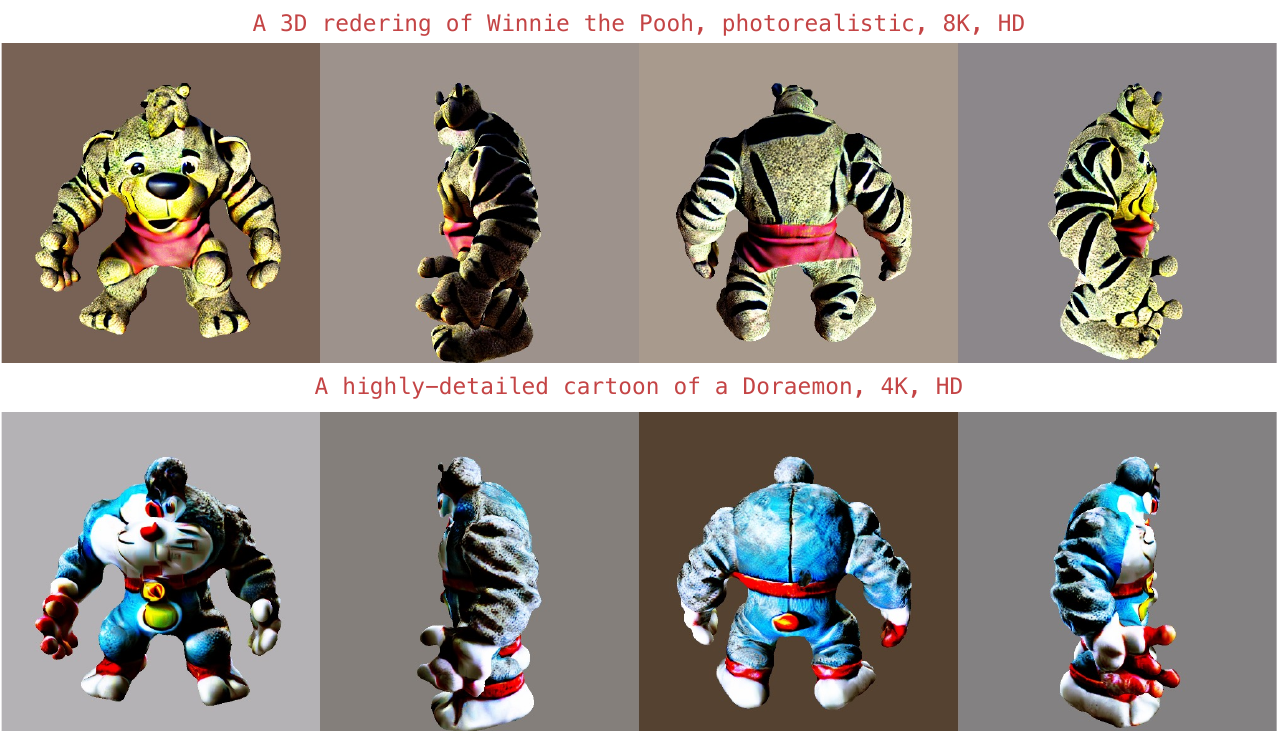} 
    \centering
    \caption{We visualize failure cases of ControlDreamer. In these examples, we use the Hulk geometry from Fig. 1 of the main manuscript. We observe that when the stylization text significantly deviates from the source geometry, it results in the generation of visually unappealing 3D models.}
    \label{fig:failure}
\end{figure*}
%%%
\begin{figure*}[t!]
\includegraphics[width=\linewidth]{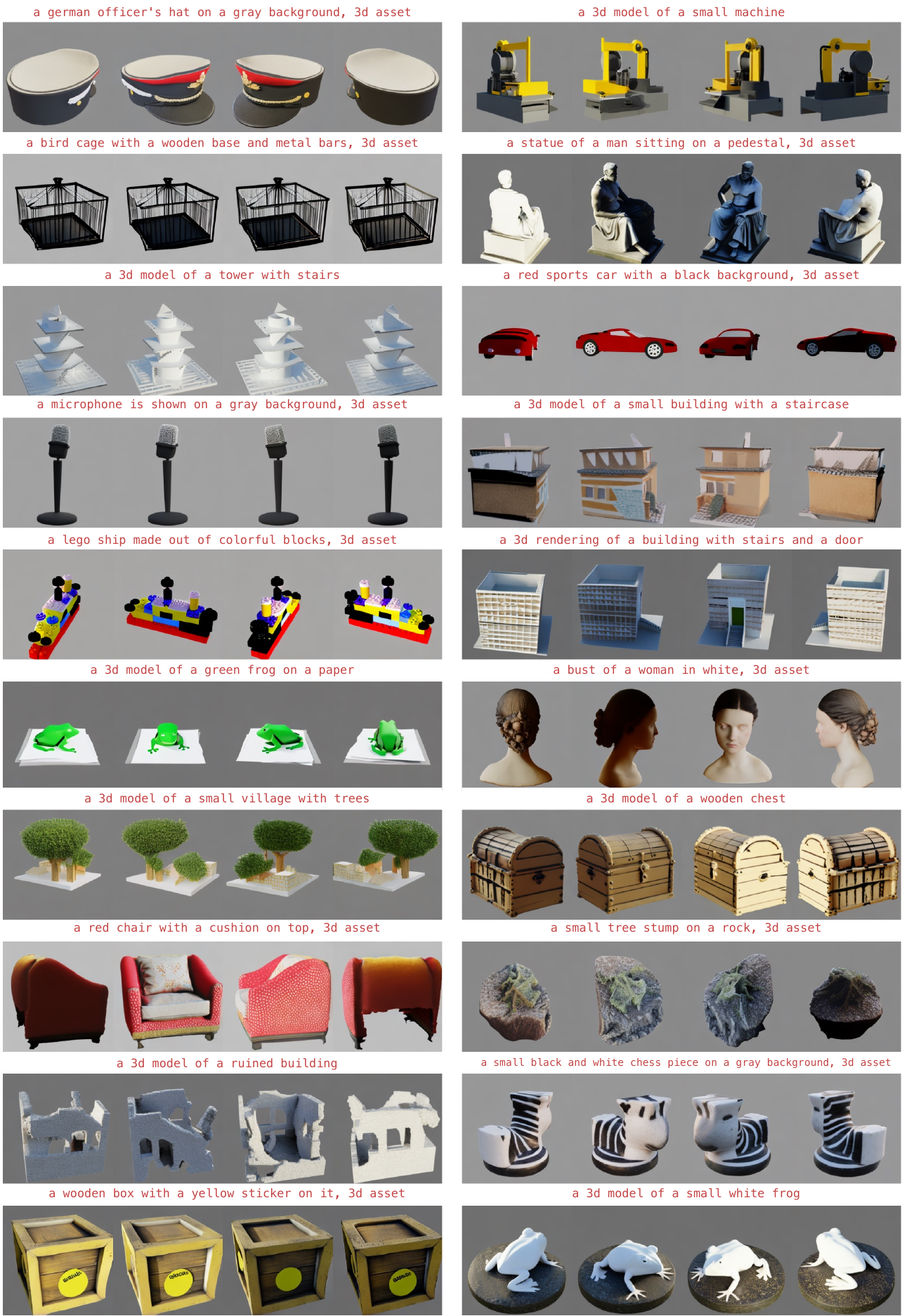} 
\caption{Examples from our generated dataset used for training MV-ControlNet, utilizing the refined text corpus. Each image was created employing a randomized camera setup.}
\label{fig:additional_1}
\end{figure*}
%%%%%%%
\begin{figure*}[t!]
\includegraphics[width=\linewidth]{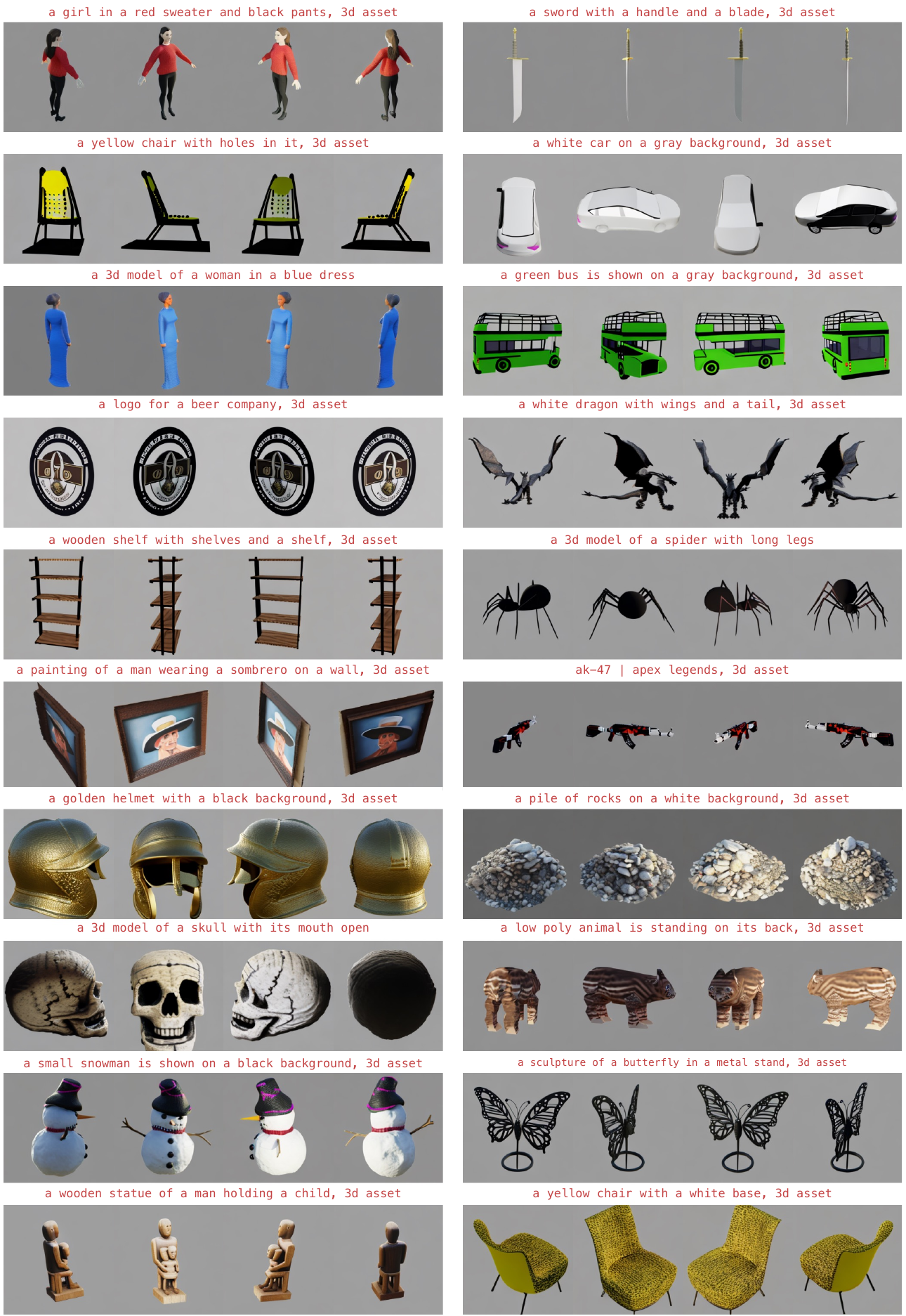} 
\caption{Additional examples of our generated dataset. A randomized camera is used for the generation of each image.}
\label{fig:additional_2}
\end{figure*}
%%%%%%%%
%%%%%%%%

\end{document}